\documentclass[
  sora,
  numberedsections,
  onecolumn
]{mirrosarticle}

\usepackage{xspace}
\usepackage[table]{xcolor}
\usepackage{amsfonts}
\usepackage{url}
\usepackage{algorithm}
\usepackage{algpseudocode}

\setlist[itemize]{leftmargin=*}
\setlist[enumerate]{leftmargin=*}

\newcommand{\method}{Code-as-World\xspace}

\definecolor{w2vblue}{HTML}{90AADC}
\definecolor{v2wred}{HTML}{E18A8A}

\newcommand{\corrauthor}{\textsuperscript{\textdagger}}

\definecolor{revisionblue}{RGB}{0, 102, 204}
\definecolor{alg_blue}{rgb}{0,0,0}

\title{Code as Worlds: \\\Large{Agentic Discovery of Executable World Representations for Physical Reasoning}}
\runningtitle{Code as Worlds}
\author{\textbf{MirroS Technical Report}}

\date{October 26, 2026}
\metadata[Blog]{\url{https://mirros.ai/blog/representing-physical-world}}
\metadata[Code]{\url{https://github.com/mirros-lab/code-as-world}}
\metadata[Project Page]{\url{https://mirros-lab.github.io/code-as-world}}

\begin{abstract}
Physical understanding and reasoning depend on forming compact and generalizable representations of the world. While modern vision-language models can recognize and explain diverse physical events, they often lack explicit representations of the underlying mechanisms—such as object states, physical parameters, and governing dynamics—needed for reliably reasoning how the world evolves and responds to interventions. In this work, we introduce \textbf{Code-as-World}, a paradigm that represents physical worlds through executable world representations. By expressing physical composition, dynamic evolution, and visual appearance as executable code, Code-as-World provides a compact, quantitatively grounded, and controllable abstraction of the physical world. To construct such representations from multimodal observations, such as natural-language descriptions or real-world videos, we develop an agentic discovery loop inspired by abductive reasoning, where an agent proposes, executes, renders, verifies, and iteratively refines executable world hypotheses. As a concrete application, we use verified executable worlds to provide scalable physical supervision for training vision-language models on quantitative physical reasoning. Experiments show that Code-as-World-VL achieves state-of-the-art performance on QuantiPhy and surpasses leading proprietary models, highlighting the potential of executable world representations as a scalable foundation for physical intelligence.
\end{abstract}

\teaserfigure[\textbf{\method} represents physical worlds as executable code for physical intelligence.\label{fig:teaser}]{%
  \includegraphics[width=\linewidth]{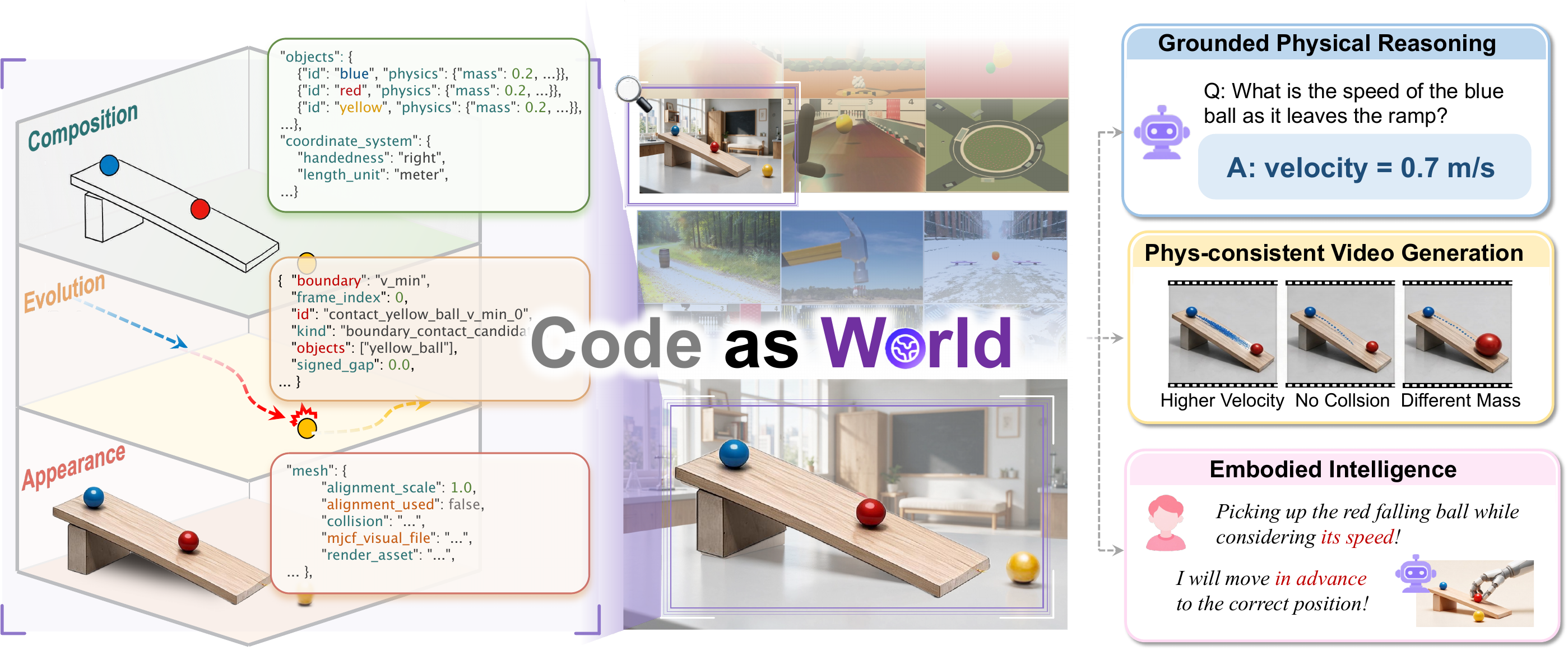}%
}

\begin{document}
\maketitle

\clearpage
\setcounter{tocdepth}{3}
\tableofcontents
\clearpage

\section{Introduction}
\label{sec:introduction}

\begin{quote}
\small
\emph{``All visible objects, man, are but as pasteboard masks.''} --- Herman Melville, \textit{Moby-Dick}, Chapter 36
\end{quote}

Understanding and reasoning about the physical world is a hallmark of intelligence \cite{bakhtin2019phyre}. Humans can make sense of unfamiliar physical situations and transfer knowledge across them, reasoning in terms of concepts such as objects, mass, motion, gravity, and friction rather than memorizing individual observations. \textit{Generalization} is therefore a defining property of physical intelligence: knowledge acquired from one situation should remain useful when objects, configurations, or observations change. Such generalization calls for \textit{representations} \cite{bengio2013representation} that compress the complexity of sensory experience into a compact abstraction of what exists in the world, how it evolves, and which factors determine its behavior.

This raises a fundamental question: what forms of representation are needed to capture, understand, and reason about the physical world? Modern vision-language models \cite{google2025gemini31flashlite,seed2026seed2, team2026kimi, qwen3_5} provide a powerful interface for describing the physical world. Trained on large-scale image–text and video–text data, they can recognize objects, narrate motion, and verbally explain a wide range of physical phenomena. Yet describing a phenomenon is not the same as recovering the mechanism that produces it \cite{machamer2000thinking}. Physical intelligence requires reasoning beyond what is visibly observed toward the structure of a system—its object states \cite{li2024core,yi2019clevrer,chow2025physbench}, physical parameters \cite{puyin2026quantiphy}, governing dynamics \cite{yi2019clevrer,sun2024probing,chow2025physbench}, and responses to interventions \cite{yi2019clevrer,luo2026vision,schulze2025visual}. This distinction gives rise to a fundamental \textit{phenomenon–mechanism dichotomy}: phenomena describe what happens, whereas mechanisms explain why it happens and predict what would happen under different conditions.

In this work, we present \textbf{Code-as-World}, a novel paradigm for representing the physical world through code. Instead of representing a world solely through pixels \cite{videoworldsimulators2024, seedance2026seedance}, latent features \cite{assran2025v, simeoni2025dinov3}, or natural-language descriptions \cite{radford2021clip, liu2023visual}, Code-as-World expresses its task-relevant structure as executable code specifying physical composition, dynamic evolution, and visual appearance. Code offers a \textit{compact yet quantitatively grounded abstraction} of the physical world: objects and their relations can be organized compositionally, while states, physical parameters, and governing dynamics remain explicitly represented. It therefore preserves the mechanistic structure needed for physical reasoning while abstracting away incidental details of individual observations.

Obtaining such a representation from incomplete observations, however, is fundamentally an inverse problem. From heliocentric theory to Newton's laws, scientific discovery has often proceeded through abductive reasoning \cite{josephson1996abductive}: recovering physical mechanisms from noisy and incomplete observations by searching for hypotheses that best explain the evidence while adhering to simplicity principles. Inspired by this process, we formulate world representation as an \textbf{agentic discovery loop} rather than a one-shot prediction problem \cite{novikov2025alphaevolve}. The \textit{executable and agent-native} nature of code enables an agent to treat each representation as a testable world hypothesis: Given multimodal evidence, such as natural-language descriptions or real videos, an agent proposes an executable world hypothesis, instantiates and executes it in a simulator, renders its predicted observations, and verifies them with the available evidence. Discrepancies are fed back to revise the hypothesis, forming a propose–instantiate–execute–render–verify loop that progressively searches for an executable world consistent with the observations.

As a concrete application, we use the resulting executable worlds as physical supervision for \textbf{quantitative physical reasoning} \cite{puyin2026quantiphy}, where a VLM must infer measurable quantities such as object size, displacement, velocity, and acceleration from monocular videos. Unlike semantic physical question answering \cite{chow2025physbench}, these tasks require the model to connect visual evidence with metric states of the underlying world. We train Code-as-World-VL, a family of vision-language models, using measurement supervision together with physical supervision derived from verified executable worlds. The resulting models substantially improve quantitative physical reasoning: Code-as-World-VL-9B outperforms substantially larger models, including Gemini-3.1-Flash \cite{google2025gemini31flashlite}, on QuantiPhy~\cite{puyin2026quantiphy}. Code-as-World-VL-27B further surpasses the 9B variant and all evaluated baselines. These results demonstrate that executable world representations can provide scalable supervision for grounding visual models in understanding physical mechanisms.

In summary, our contributions are as follows:

\begin{itemize}
    \item \textbf{Executable world representation.} Based on a dedicated discussion among existing physical world representations, we introduce Code-as-World, which represents task-relevant physical worlds as executable code over physical composition, dynamic evolution, and visual appearance.
    \item \textbf{Agentic discovery loop.} We formulate the world representation process as an agentic discovery problem, and develop a propose–instantiate–execute–render–verify loop that iteratively searches for world representations consistent with language or visual evidence.
    \item \textbf{Physical supervision for VLMs.} We use verified executable worlds to provide scalable supervision for quantitative physical reasoning. Code-as-World-VL achieves great improvements and attains state-of-the-art performance on QuantiPhy, outperforming leading proprietary models.
\end{itemize}
\section{In Search of Physical World Representations}

Physical understanding \cite{yi2019clevrer, chow2025physbench, zhou2025paibenchcomprehensivebenchmarkphysical, bakhtin2019phyre, puyin2026quantiphy} ultimately depends on the representation through which a model encodes the world. Existing approaches have explored this question from several perspectives, including pixel-level prediction, geometric reconstruction, and language-based abstraction. These paradigms have each captured important aspects of the physical world, yet none alone provides a complete representation that simultaneously supports semantic understanding, structural composition, and temporal prediction.

\paragraph{Pixels.}
A natural approach to modeling the physical world is to learn directly from visual observations. Video generative models learn temporal dynamics by predicting future observations from past frames, achieving increasingly realistic and diverse generations \cite{videoworldsimulators2024,seedance2026seedance,wu2024ivideogpt,huang2025vid2world,chen2026harnesseval}. However, observation prediction alone does not require a model to explicitly represent the underlying causes of visual changes. If a model only optimizes future pixel prediction, it does not need to distinguish whether a scene change is caused by camera motion or object motion, nor whether an object is temporarily occluded or has disappeared. Multiple contradictory internal explanations may therefore lead to similar predictive accuracy as long as they produce visually plausible futures.

This ambiguity limits physical reasoning. A physically meaningful representation must disentangle the factors that generate observations, including persistent object identity, physical state transitions, interactions, and viewpoint changes. A visually plausible future is not necessarily a physically correct future \cite{kang2024far, motamed2026generative}.

\paragraph{3D.}
Another line of research focuses on reconstruction, including 3D reconstruction \cite{mildenhall2021nerf,kerbl3Dgaussians, dust3r_cvpr24, wang2025vggt}, inverse graphics \cite{boss2021nerd,gao2024relightable,jiang2024gaussianshader,NeRFactor}, and dynamic scene representations \cite{pumarola2020dnerf,park2021hypernerf,yang2024deformable,wu20244d,yang2023gs4d,duan:2024:4drotorgs}. By requiring models to recover geometry, viewpoint, and appearance, reconstruction-based approaches provide strong constraints on information preservation and achieve impressive capabilities in scene modeling.

However, reconstructability does not necessarily imply interpretability. A latent representation may faithfully reproduce an observation while entangling persistent structures, dynamic variables, and appearance factors. Recovering the 3D geometry of an object does not explain its physical behavior, such as why it falls after support is removed or how it interacts with other objects. Reconstruction preserves information, but does not automatically uncover the causal factors that govern the evolution of the world.

\paragraph{Natural language.}
Language provides another powerful abstraction of the physical world \cite{berg2025semantic, chen2025planning, vista2026}. By converting visual observations into captions, descriptions, or reasoning traces, language-based representations capture high-level concepts that are compact, compositional, and transferable across contexts to support reasoning. For example, a description such as ``a person picks up a cup" preserves entities, actions, and semantic relations while discarding irrelevant visual details.

However, language is inherently limited in expressing the continuous and quantitative aspects of physical states. Precise geometry, trajectories, contact relationships, and physical parameters are difficult to encode precisely through a small number of discrete tokens. Language provides an effective semantic interface, but not a complete physical state representation.

These perspectives reveal complementary strengths \cite{wu2026visual}: pixels preserve rich details, 3D representations preserve geometric structure, and language captures semantic abstractions. Yet physical intelligence may require a representation that better integrates these capabilities: semantic like language, structured like reconstruction, and capable of modeling temporal evolution like generative models. Such a representation should not merely describe or reproduce the world, but explicitly capture the entities, states, and mechanisms that generate observations.

\begin{figure}[t]
    \centering
    \includegraphics[width=0.99\linewidth]{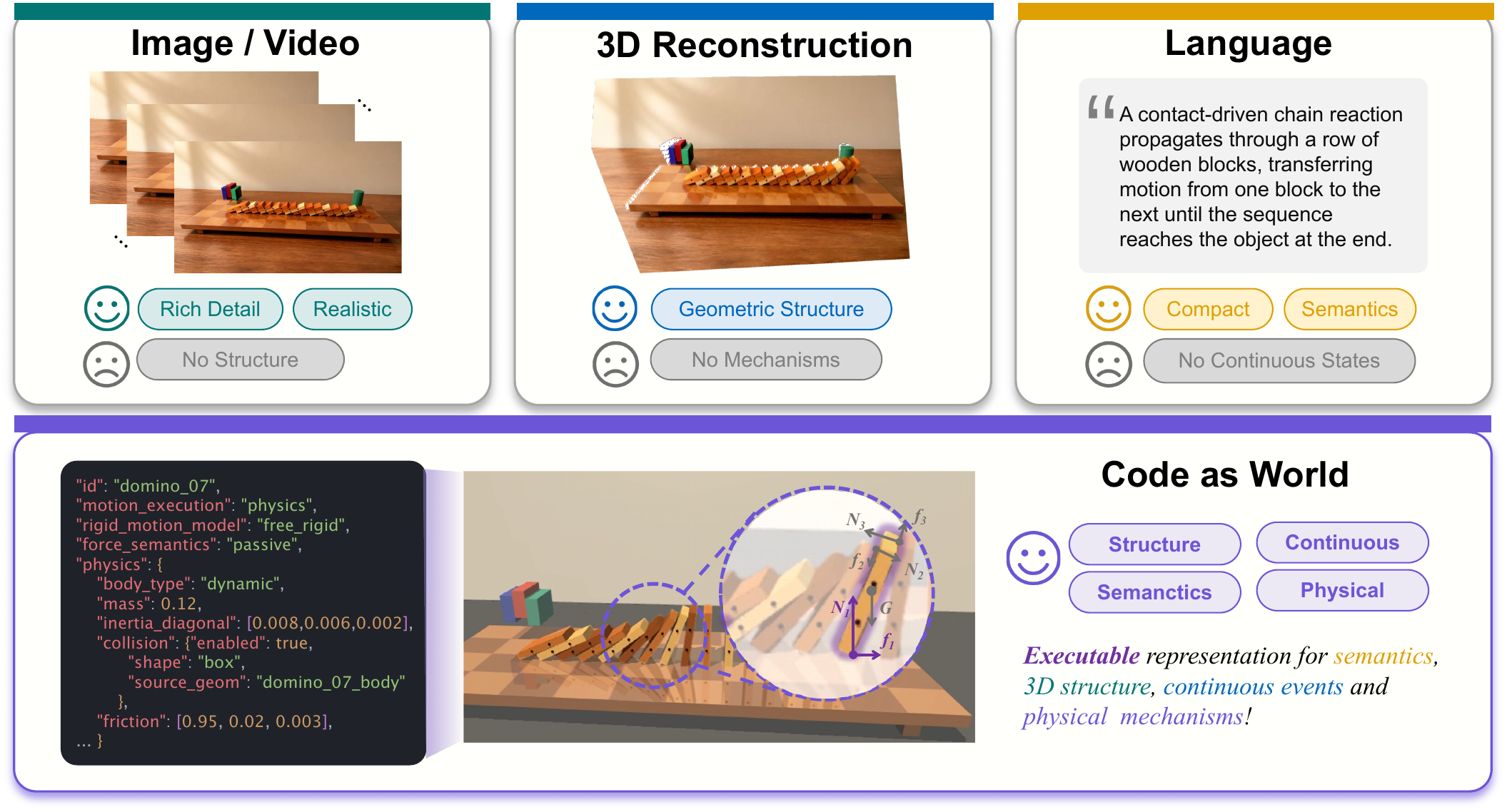}
    \caption{
    \textbf{Comparison among different physical world representations.}
Pixels preserve rich visual detail but lack explicit structure, leaving the underlying causes of observations ambiguous; 3D representations capture geometric structure but do not necessarily reveal physical mechanisms; and language provides compact semantic abstractions but lacks precision for continuous physical states. Code complements these representations by expressing the world as an abstract, quantitatively grounded, and executable representation.
    }
    \label{fig:representation}
\end{figure}

\section{Code as Worlds: Executable World Representations}
\label{sec:world-representation}

In this work, we introduce \method, a paradigm that represents the physical world through code.

\paragraph{Code} provides a structured and compositional abstraction of the world: it explicitly represents entities, relations, physical parameters, and events as operations over states, while abstracting away incidental details of individual observations. Such a representation separates the structured state underlying physical processes from the continuous appearance through which they are observed. The former provides compositionality, editability, and explicit constraints for reasoning, while the latter preserves rich visual details and uncertainty that are difficult to explicitly encode. Through execution and rendering, code can generate observations consistent with the represented world, where physical equivalence—the consistency of world composition, constraints, and evolution—is prioritized over pixel-level duplication.

\subsection{Practical Implementation}

To instantiate this paradigm, we develop a concrete implementation of executable world representation (EWR) coupled with programmatic calls to a physical simulation engine. Figure~\ref{fig:scene-representation} presents an example of the representation and its simulation interface.

Specifically, an EWR $p$ conceptually consists of three components: 
\begin{equation}
  p=\left(\mathcal C,\mathcal E,\mathcal A\right)
  \in\mathcal P_{\mathrm{exec}},
  \label{eq:world-program}
\end{equation}
where $\mathcal P_{\mathrm{exec}}$ denotes the space of valid executable worlds.
Its three components jointly specify what the world contains, how it evolves, and how it appears.

\paragraph{Physical composition.}
$\mathcal C$ describes what exists in a world and the relatively stable physical properties of its entities. It includes objects in the scene, their geometry, metric dimensions, and physical properties such as mass, friction, and gravity. Environmental structures such as floors, tables, and walls are likewise represented as static physical entities, allowing them to participate in support, contact, and collision. This component defines the persistent entities involved in physical processes and the fundamental conditions under which they unfold.

\paragraph{Dynamic evolution.}
$\mathcal E$ describes how the world unfolds over time. It includes initial object states, their temporal changes, key events, and the duration of the simulation. Given the dynamic evolution component, the world composition can be expanded into a complete state trajectory, producing events such as contacts, collisions, velocity changes, and termination conditions during execution.

\paragraph{Visual appearance.}
$\mathcal A$ describes how the physical world is observed and presented. It includes camera parameters, backgrounds, materials, lighting, output frame rates, resolutions, and rendering or video-generation configurations. This component does not alter the underlying physical process but determines how the physical trajectory is visually presented. Environmental elements that participate in support or collision belong to physical composition, whereas backgrounds and appearance factors that provide only visual context belong to visual appearance.

This implementation exposes the represented world through an executable and controllable interface. Different components of an EWR can be inspected, modified, and executed independently: an object, physical parameter, dynamic condition, or camera setting can be changed while preserving the remaining structure. Such executable representations provide a foundation for downstream reasoning, simulation, verification, and data generation.

\section{Agentic Discovery of World Representations}
\label{sec:data-engine}

While Code-as-World provides a structured representation space for physical worlds, recovering such representations from partial and heterogeneous observations remains challenging. To address this challenge, Code-as-World formulates world representation as an agentic discovery process rather than a direct prediction problem. Given multimodal evidence, including text descriptions and real videos, the discovery agent constructs semantic or visual constraints and iteratively searches for an EWR consistent with the input evidence through a shared \emph{propose--instantiate--execute--render--verify} loop. The resulting representation externalizes the underlying physical mechanism as an interface that an agent can directly query, intervene on, and verify, supporting physical data generation, quantitative supervision, verifiable reward construction, and broader downstream physical reasoning.

\subsection{Evidence Construction}
\label{sec:modality-specific-processing}

Given an input $\xi$ from modality $m$, the agent first constructs modality-specific evidence $\eta$ through a dedicated evidence adapter before entering the discovery loop. Text and video inputs are processed by different adapters, which transform them into semantic or visual evidence that constrains the same EWR space.

\paragraph{Text-driven world construction.} The agent extracts explicit entities, spatial relations, physical events, and intended outcomes from the input text and organizes them as structured semantic evidence. Because textual descriptions rarely determine geometry, physical parameters, or camera configurations completely, the agent combines physical priors with reasonable default conditions to initialize an executable world hypothesis, which is progressively refined through simulation and semantic verification. For downstream applications as in Section~\ref{sec:training}, once the executable world is obtained, we apply a video generation model for sim-to-real transfer, enriching objects, materials, backgrounds, and lighting. The resulting videos achieve greater visual realism while remaining aligned with the underlying physical trajectories and world states.

\paragraph{Video-driven world abstraction.}
Given a real video, the agent extracts depth maps, instance masks, and object tracks as visual evidence for world construction and verification. Depth maps constrain the scene’s spatial structure and relative distances, masks identify object boundaries and geometric extent, and tracks capture temporal correspondences and image-plane motion. For each segmented scene object, it further employs a 3D object generation model~\citep{chen2025sam3d} to construct a corresponding mesh. It then combines these meshes with depth and tracking evidence to recover the objects' spatial positions, scales, and dynamic states, thereby constructing an executable 3D scene. The rollout of each candidate EWR is projected back into the input view and iteratively refined by evaluating its consistency with the observed geometry, depth, masks, and trajectories.

\begin{figure*}
  \centering
  \includegraphics[width=\linewidth]{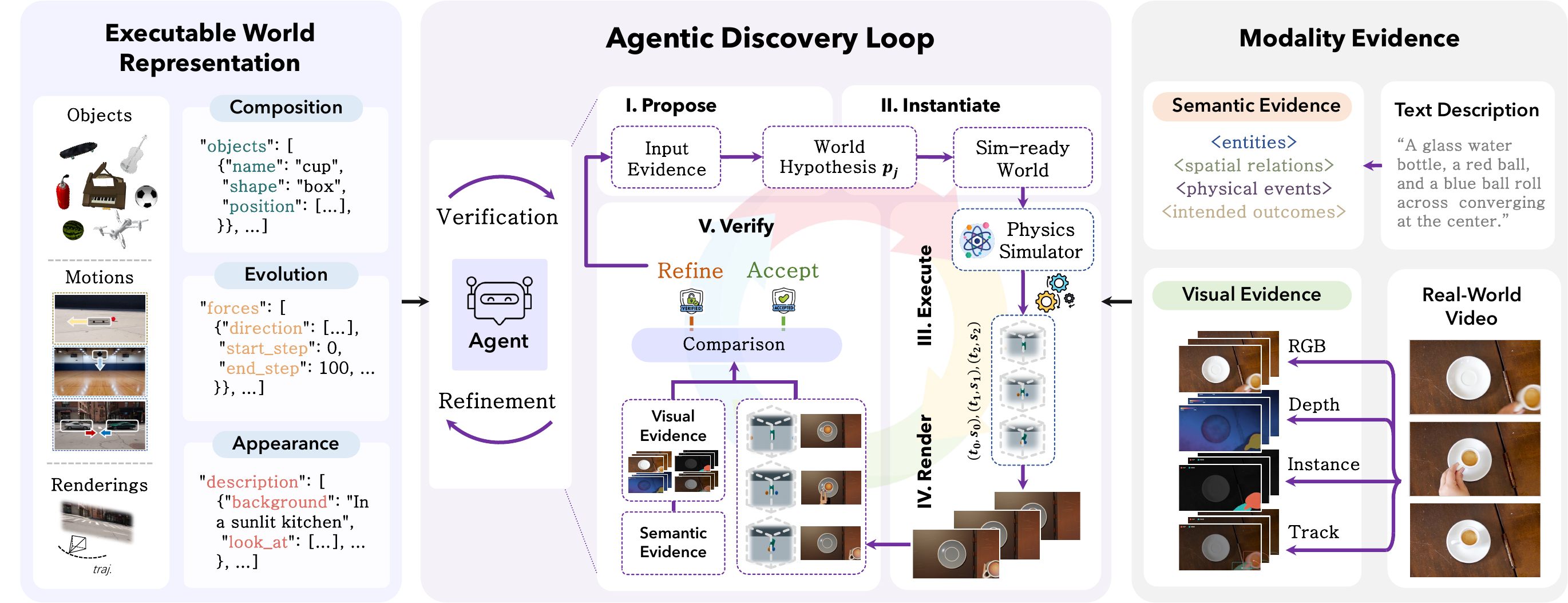}
  \caption{\textbf{Agentic discovery loop of executable world representations.}
Given a text prompt or real video, modality-specific processors transform the input into semantic or visual evidence.
A shared propose--instantiate--execute--render--verify loop then optimizes an executable world representation over physical composition, dynamic evolution, and visual appearance.
}
  \label{fig:pipeline}
\end{figure*}

\subsection{Agentic Discovery Loop}
\label{sec:world-hypothesis-optimization}

Given modality-specific evidence $\eta$, the discovery agent proposes an initial EWR and progressively refines it through multiple iterations of the \emph{propose--instantiate--execute--render--verify} loop, searching the executable hypothesis space for an EWR that best explains the input evidence while remaining as parsimonious as possible. Algorithm~\ref{alg:agentic-discovery} summarizes the whole procedure of the agentic discovery loop.

At each iteration, the agent proposes or updates an EWR $p=(\mathcal C,\mathcal E,\mathcal A)$ based on $\eta$, the current hypothesis, and structured feedback $\Delta$ from the previous iteration. The EWR is then instantiated as simulator-ready parameters $\theta$ that conform to a specific simulator interface, which the simulator executes to produce a complete state trajectory $\tau$. This trajectory explicitly records object states, contacts, collisions, and event outcomes, providing the world hypothesis with temporal and causal consequences that can be directly inspected.

The agent subsequently renders $\tau$ into predicted visual observations. For video input, it additionally projects the simulated states into depth maps, instance masks, and image-plane trajectories. During verification, the predicted and input evidence are compared at selected key frames. For text input, the verifier primarily evaluates semantic and physical constraints; for video input, it jointly compares RGB appearance, depth, masks, and trajectories. Frame-level discrepancies are aggregated into $\Delta$, which guides the agent in locally revising the relevant component in the next iteration. The loop terminates when the current EWR explains the input sufficiently well and parsimoniously. Otherwise, refinement continues until the iteration budget is exhausted, at which point the hypothesis is rejected.

\begin{algorithm}[t]
\caption{Agentic Discovery of Executable World Representations}
\label{alg:agentic-discovery}
\begin{algorithmic}[1]
\Require Input evidence $\xi$; {\color{alg_blue} LLM $A$}; iteration budget $K$

\Statex \parbox[t]{0.94\linewidth}{
\textbf{Notation.} $\eta$ denotes the preprocessed evidence.
The current EWR is $p=(\mathcal C,\mathcal E,\mathcal A)$, while $\theta$
denotes its simulator-ready instantiation and $\tau$ its executed
world-state trajectory. The structured trace $z$ records the hypotheses
and verification outcomes produced during discovery.
}

\Ensure An evidence-supported EWR $p$, or rejection, together with trace $z$

\If{$\xi$ is a text specification}
    \State $m\gets\mathrm{text};\quad
        \eta\gets\operatorname{InterpretText}({\color{alg_blue}A},\xi)$
        \Comment{Semantic evidence}
\ElsIf{$\xi$ is a video}
    \State $m\gets\mathrm{video};\quad D\gets\operatorname{Depth}(\xi); \quad M\gets\operatorname{Segment}(\xi);\quad Q\gets\operatorname{Track}(\xi)$
        \Comment{Visual evidence}
    \State $\eta\gets(\xi,D,M,Q)$
\EndIf
\State $p\gets\varnothing;\quad\Delta\gets\varnothing;\quad z\gets\varnothing$

\For{$k=1$ to $K$}
    \State $p\gets\operatorname{ModifyEWR}({\color{alg_blue}A}, \eta,p,\Delta)$
        \Comment{\textsc{Propose | Update}}

    \State $\theta\gets\operatorname{CompileEWR}(p)$
        \Comment{\textsc{Instantiate}}

    \State $\tau\gets\operatorname{RunSimulation}(\theta)$
        \Comment{\textsc{Execute}}

    \State $\hat X\gets\operatorname{Render}(\tau,\theta); \quad (\hat D,\hat M,\hat Q)\gets\operatorname{Project}(\tau,\theta)$ \Comment{\textsc{Render \& Project}}
    \State $\hat\eta\gets(\hat X,\hat D,\hat M,\hat Q)$

    \State $F\gets
        \operatorname{SelectFrames}(m,\eta);\quad
        \Delta\gets\varnothing$
        \Comment{\textsc{Verify}}

    \ForAll{$f\in F$}
        \State $\delta\gets\operatorname{CompareAndDiagnose}({\color{alg_blue}A},\hat \eta[f],\eta)$
        \State $\Delta\gets\Delta\oplus\delta$
    \EndFor
    
    \State $z\gets z\oplus\operatorname{RecordRound}(k,p,\tau,\Delta)$
    \If{$\operatorname{Accept}({\color{alg_blue}A},p,\Delta)$}
        \State \Return $(p, z)$
    \EndIf
\EndFor

\State \Return $(\textsc{Reject},z)$
\end{algorithmic}
\end{algorithm}

\subsection{Evaluation}
\label{sec:experiments}
\subsubsection{Experimental Setup}
\label{sec:engine-setup}
\paragraph{Data Sources and Filtering.}
For text-driven construction, language specifications are generated by LLMs and subsequently reviewed by human annotators. For video-driven abstraction, candidate observations are selected from WISA-80K~\citep{wang2026wisa} through a motion-focused filtering pipeline. We first retain clips whose metadata indicates salient rigid-body motion or collision-like interactions and remove clips that mix unrelated physical phenomena. We then reject videos with substantial camera translation or rotation, insufficient object motion, severe editing, or incomplete physical events. The remaining clips undergo manual review for temporal continuity, object visibility, and suitability for executable reconstruction.

\paragraph{Processing Tools.}
For each retained video, SAM3~\citep{carion2025sam3} supplies instance masks and image-plane tracks, VGGT-Omega~\citep{wang2026vggt} estimates scene depth and camera geometry, and SAM3D~\citep{chen2025sam3d} provides object geometry. These observations constrain the agentic discovery loop, and a resulting world is retained only if it can be rendered and verified against the source video. Across both input modalities, \method uses the same simulation interface to produce synchronized videos and physical states.

\paragraph{Evaluation Protocol and Metrics.}
We evaluate video-driven world reconstruction for up to $K=5$ agentic discovery rounds and compare iterative refinement with one-shot generation and Best-of-$5$ sampling under matched evaluation budgets. Visual Alignment and Object IoU~\citep{perazzi2016benchmark} measure full-video visual agreement and object-region overlap between an input video and its simulator reconstruction. Motivated by~\citep{gupta2018social,jiang2021cotr}, we use Traj-ADE, Velocity-ADE, and Accuracy@$2\%D$ to measure object-position error, inter-frame displacement error, and the percentage of valid observations within $0.02D$, respectively, with distances normalized by the frame diagonal $D$.

\subsubsection{Qualitative Analysis}

\begin{figure}[p]
    \centering
    \includegraphics[
        width=\textwidth,
        keepaspectratio
    ]{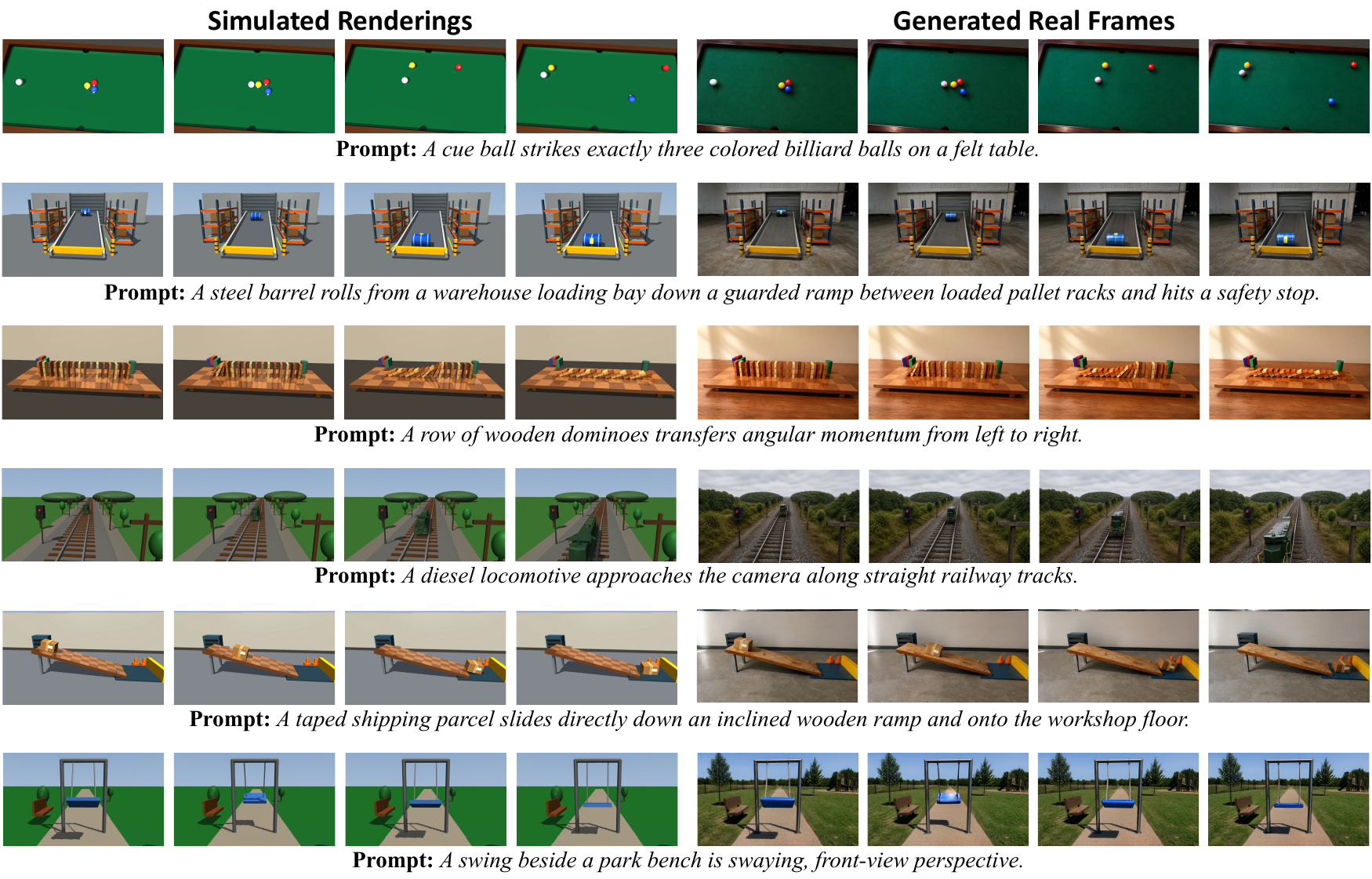}
    \caption{\textbf{Visual results of text-driven construction}: from executable simulation to realistic video. For each example, the left side shows matched frames from a simulator rollout generated from a text specification, while the right side shows the corresponding final video synthesized by the video generation model. The sim-to-real transformation introduces rich and diverse objects, materials, backgrounds, and textures while preserving the motion trajectories and physical evolution encoded by the executable world.}
    \label{fig:text-driven-examples}
\end{figure}

\begin{figure}[p]
    \centering
    \includegraphics[
        width=\textwidth,
        keepaspectratio
    ]{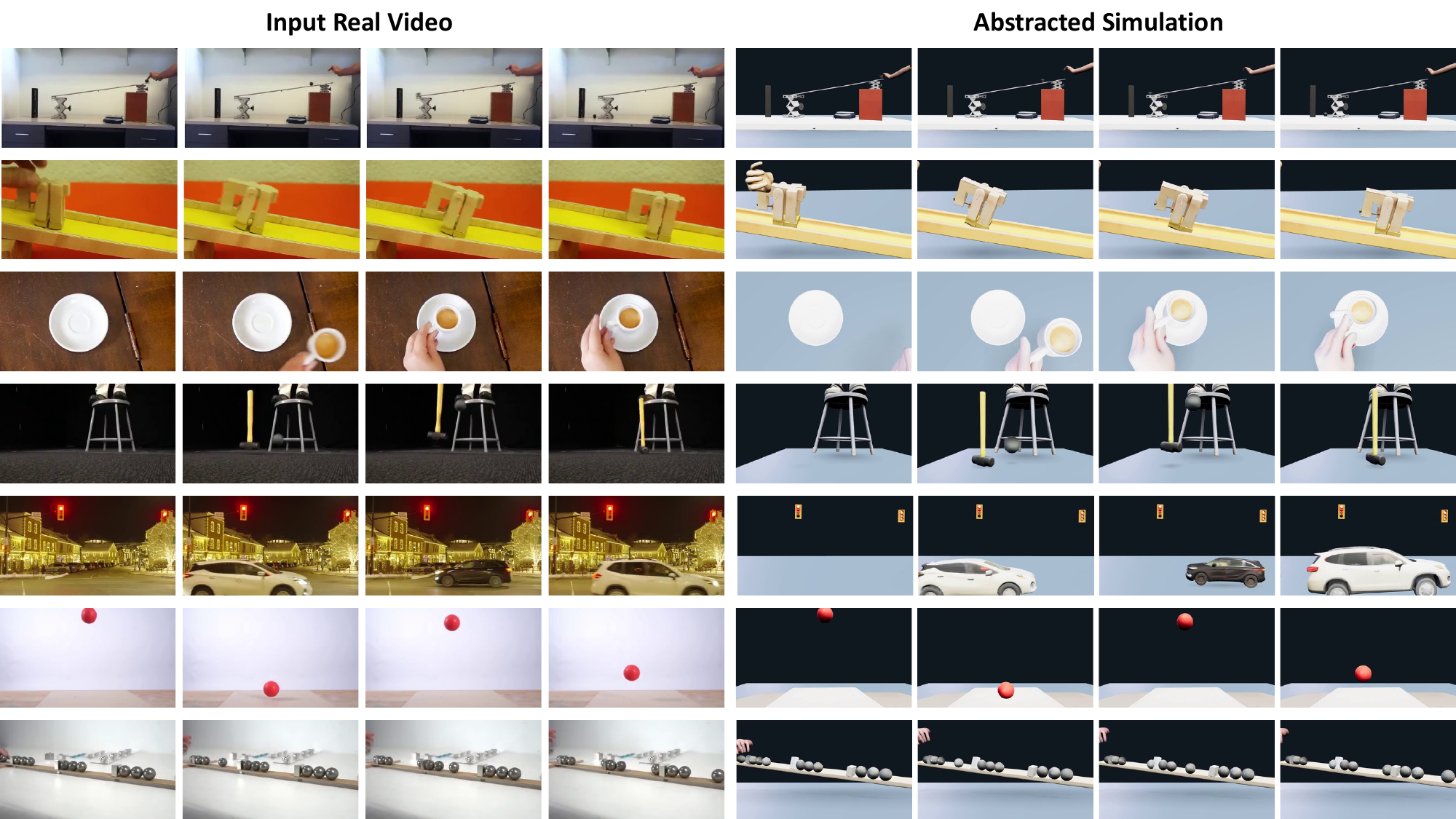}
    \caption{\textbf{Visual results of video-driven abstraction}: from real video to executable simulation. For each example, the left side shows matched frames from the input real video, while the right side shows the corresponding rollout of the abstracted executable world. The recovered simulation aligns with the observed scene composition, object motion, and physical interactions, capturing the underlying physical mechanism as a coherent, executable process.}
    \label{fig:video-driven-examples}
\end{figure}

We evaluate the quality of data curated by \method{} from both text and video inputs. Figures~\ref{fig:text-driven-examples}
and~\ref{fig:video-driven-examples} visualize representative inputs and
outputs for the two modalities. These examples demonstrate that \method{} can both turn semantic descriptions into physically grounded videos and recover executable worlds from authentic visual observations.

\begin{figure}[t]
    \centering
    \includegraphics[
        width=\textwidth,
        height=0.34\textheight,
        keepaspectratio
    ]{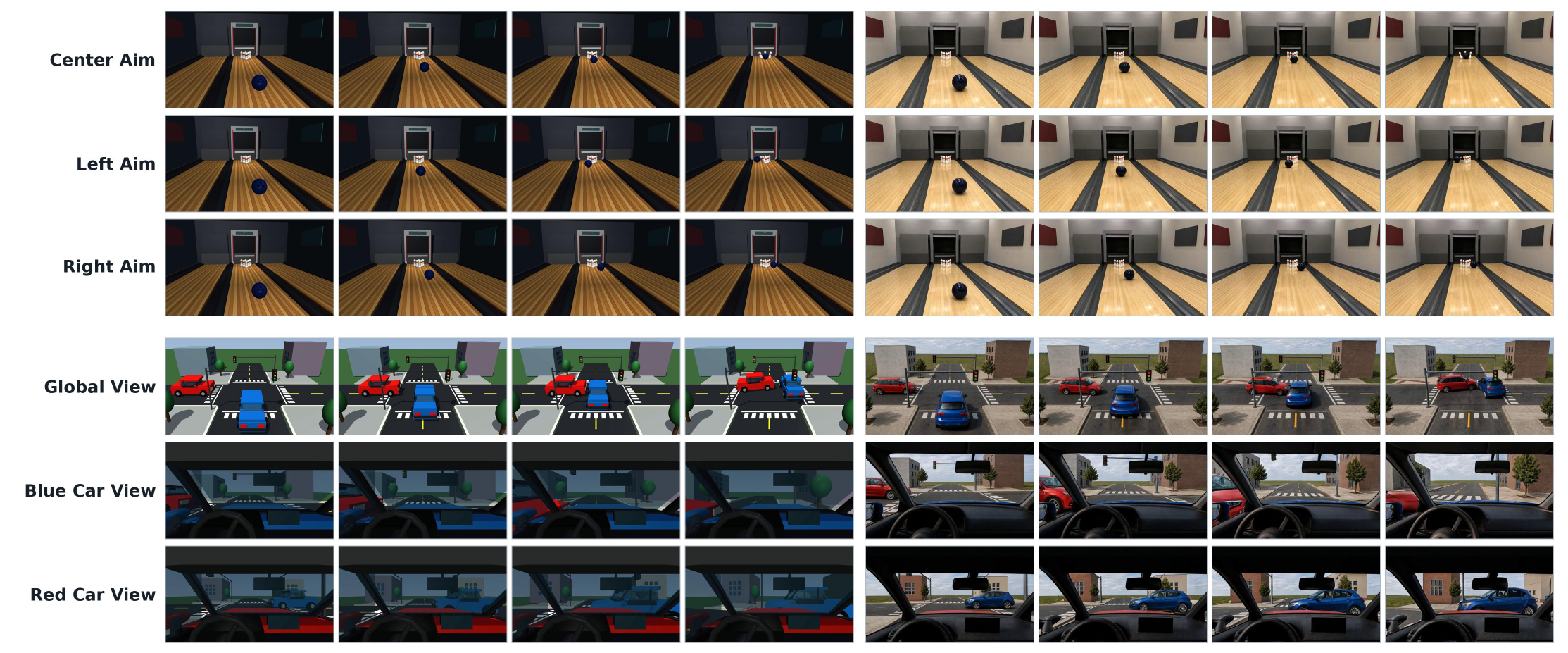}
    \caption{\textbf{Controllable resimulation with \method{}.}
    In each row, the left four columns show an edited simulator rollout, and
    the right four show temporally aligned frames from its realistic video.}
    \label{fig:controllable-resimulation}
\end{figure}

Beyond the observed evidence, \method{} exposes recovered executable worlds
as editable programs for controllable generation. As shown in
Figure~\ref{fig:controllable-resimulation}, we can resimulate a world after changing physical quantities or initial conditions. For example, changing the bowling ball's initial velocity direction produces distinct trajectories. We can also change the camera configuration to render the same collision from a global view or from either car. Our internal video generation model renders each edited rollout as a realistic video while preserving the specified physical evolution and viewpoint. Separating world editing from appearance synthesis enables coherent counterfactual videos without
reconstructing every variant.

\subsubsection{Agentic Discovery Loop}
\label{sec:data-curation-quality}

\begin{figure}[t]
    \centering
    \includegraphics[width=\textwidth]{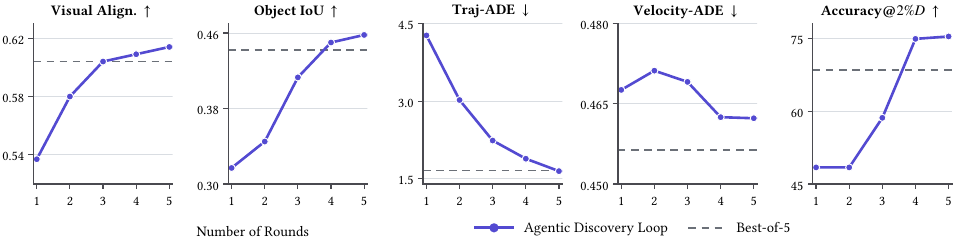}
    \caption{\textbf{Agentic discovery across loop rounds.} Solid curves connect one-shot to agentic loop rounds two through five. Dashed gray lines mark Best-of-$5$ from independent sampling. At a matched five-evaluation budget, agentic loop outperforms on Visual Alignment, Object IoU, Traj-ADE, and Accuracy@$2\%D$. Here, $D$ denotes the frame diagonal. Traj-ADE is reported in $\%D$, Velocity-ADE in $\%D$/step, and accuracy at $2\%D$.}
    \label{fig:real2sim_agentic_ablation}
\end{figure}

To evaluate the effectiveness of the agentic discovery loop, we set the maximum number of loop rounds to $K=5$ and compare different rounds for video-driven world
reconstruction. Candidate selection and refinement are driven by the verifier in \method{}, whereas the results are measured using independent metrics that are not included in the verification signal. All reported metrics compare the input video with the simulator rendering of the reconstructed EWR. Figure~\ref{fig:real2sim_agentic_ablation} plots each metric as a function of
the loop rounds. Each solid curve begins with the one-shot result at
round one and continues with four successive agentic discovery loop updates
at rounds two through five. The dashed gray line marks the Best-of-$5$
reference. Across these refinements, static quality and most motion-fidelity
metrics improve overall. At the matched five-evaluation budget, the agentic
discovery loop outperforms Best-of-$5$ on most aspects, demonstrating more effective use of compute than independent sampling.

\subsection{Limitations}

Real-world environments are highly complex, and current simulators cannot faithfully capture the full range of physical conditions. Even seemingly simple rigid-body motion can be sensitive to small variations in terrain, contact geometry, material properties, or other latent factors. When the underlying process falls outside the simulator's modeling scope, the agentic discovery loop may converge to a locally plausible EWR without recovering a mechanistically accurate explanation of the physical process.

\section{Application: Learning Quantitative Physical Reasoning}
\label{sec:training}

As a concrete downstream application of Code-as-World, we study quantitative physical reasoning in vision-language models \citep{puyin2026quantiphy}. Unlike semantic physical question answering, which can often be solved through visual patterns or commonsense priors, this task requires a VLM to infer physical quantities hidden beneath pixel observations in monocular videos, such as an object's real-world size, velocity, and acceleration. However, real-world videos rarely provide annotations of such underlying physical quantities. Code-as-World addresses this fundamental challenge by providing verified executable worlds that expose physical states and trajectories, enabling scalable physical supervision.

\subsection{Problem Formulation}
\label{sec:problem}

Given a monocular video $V$ and a quantitative physical question $q$, the model predicts a numerical answer $\hat y=f_\theta(V,q)\in\mathbb R$. The question specifies a target object, relevant timestamps, a queried quantity---such as size, velocity, or acceleration---and an output unit in either pixel space or world space. For world-space queries, the question additionally provides a reference quantity with a known world-space value $\rho$ to enable metric calibration.

The video provides measurements in pixel space but does not reveal their world-space scale. Let $y^{\mathrm{pix}}$ and $\rho^{\mathrm{pix}}$ denote the target and reference measurements in pixel units. Following QuantiPhy~\citep{puyin2026quantiphy}, the relative scale is estimated as $\gamma=\frac{\rho}{\rho^{\mathrm{pix}}}$ and $y=\gamma y^{\mathrm{pix}}$ is the ground-truth answer of the world-space quantity. The same calibration converts pixel measurements into world units for size, displacement, velocity, and acceleration. For 3D settings, depth information provides additional geometric cues that help relate image-space measurements to their corresponding world-space quantities.

\subsection{Image-Space Measurement Grounding}
\label{sec:pixel-grounding}

Quantitative physical reasoning requires reliable image-space measurements as a foundation \cite{yu2016modeling, peng2023kosmos}. 
We construct pixel-level supervision by converting bounding boxes, masks, and object tracks from existing visual datasets into quantitative question-answer pairs. These questions involve object extent, position, displacement, velocity, and acceleration, and require no world-space calibration. Specifically, for an object trajectory $\{\mathbf c_t\}_{t=1}^{T}$ sampled at interval $\Delta t$, displacement, velocity, and acceleration are computed as
\begin{equation}
    \mathbf d
    =\mathbf c_{t_2}-\mathbf c_{t_1},
    \qquad
    \mathbf v_t
    =\frac{\mathbf c_{t+1}-\mathbf c_{t-1}}{2\Delta t},
    \qquad
    \mathbf a_t
    =\frac{\mathbf c_{t+1}-2\mathbf c_t+\mathbf c_{t-1}}
           {\Delta t^2},
    \label{eq:pixel-motion}
\end{equation}
where $t_1$ and $t_2$ are the timestamps specified by a displacement question. Object extent and position are read directly from bounding boxes or masks, while motion quantities are derived from tracks.

We train the model on the resulting pixel-level dataset $\mathcal D_{\mathrm{pix}}$ with supervised fine-tuning:
\begin{equation}
    \mathcal L_{\mathrm{pix}}
    =
    -\mathbb E_{(V,q,y)\sim\mathcal D_{\mathrm{pix}}}
    \log \pi_\theta(y\mid V,q).
    \label{eq:pixel-loss}
\end{equation}
This stage teaches the model to localize, measure, and track objects, providing the visual foundation for subsequent world-space physical reasoning.

\subsection{World-Space Physical Calibration from Verified Executable Worlds}
\label{sec:executable-rl}

After image-space grounding, we further use verified executable worlds to construct physical supervision. Each EWR provides a synchronized video and simulated state trajectory that records object geometry, camera parameters, timestamps, and time-varying physical states. This enables direct generation of quantitative question-answer pairs with exact world-space labels. Specifically, we sample a target object, relevant timestamps, a physical quantity, and a requested world unit from this record, optionally providing a reference quantity with a known world-space value as a scale prior. The answer $y$ is obtained directly from the same EWR: object size is read from the scene geometry, while displacement, velocity, and acceleration are computed from the state trajectory. A question is retained only when its target object, temporal range, and optional reference are valid in the corresponding observation, yielding a VQA training instance $(V,q,y)$. Text-driven and video-driven executable worlds use the same format, forming $\mathcal D_{\mathrm{text}}$ and $\mathcal D_{\mathrm{video}}$, respectively, which we combine into unified world-level training data.

We optimize the model on these executable-world examples using Group Relative Policy Optimization
(GRPO)~\citep{shao2024deepseekmath,guo2025deepseek}. The reward combines scale-normalized numerical accuracy with auxiliary rewards for unit correctness and response format:
\begin{equation}
    r_{\mathrm{num}}
    =
    \exp\left(
        -\frac{|\hat y-y|}{|y|+\epsilon}
    \right),
    \qquad
    r
    =
    r_{\mathrm{num}}
    +\lambda_u r_{\mathrm{unit}}
    +\lambda_f r_{\mathrm{fmt}}.
    \label{eq:numerical-reward}
\end{equation}

The two sources of executable-world supervision provide complementary benefits. Text-driven worlds offer fully observable simulator states and numerically exact physical supervision, while video-driven worlds better match the appearance and motion distributions of real observations. Joint training therefore combines accurate physical supervision with visual generalization to real-world videos.

\begin{figure}[t]
    \centering
    \includegraphics[width=\linewidth]{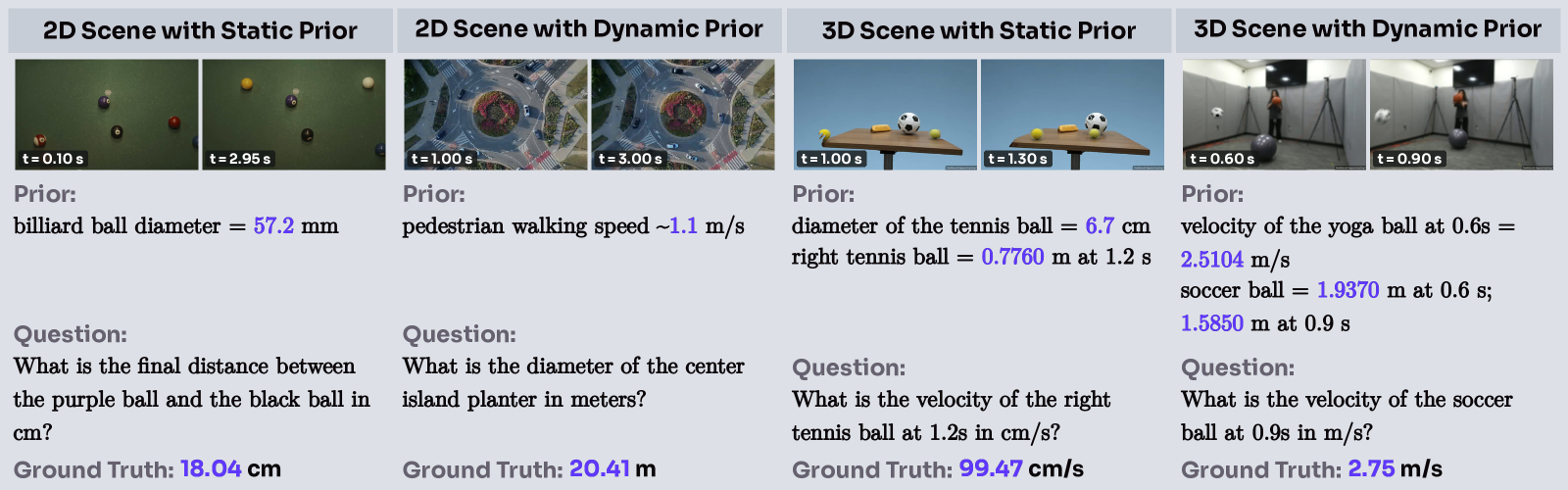}
    \caption{\textbf{Illustration of quantitative physical reasoning.} A video and a question are given as input. For 3D scenes, an additional depth prior provides spatial context. The model needs to predict an answer grounded in the video. Examples adapted from~\citep{puyin2026quantiphy}.}
    \label{fig:vqa-example}
\end{figure}

\begin{table}[p]
  \centering
  \scriptsize
  \definecolor{ourspurple}{HTML}{EAE9F9}
  \definecolor{groupgray}{RGB}{245,245,245}
  \caption{\textbf{Quantitative physical reasoning results.} We report MRA on the 2S, 2D, 3S, and 3D subsets and their macro-average on QuantiPhy-validation. The 4B and 9B variants produce direct answers, whereas the 27B reasoning variant is scored only on the answer emitted after its reasoning trace. Rows are grouped by model family, and \method{}-VL variants are highlighted in purple.}
  \label{tab:quantiphy-validation}
  \setlength{\tabcolsep}{10pt}
  \resizebox{\textwidth}{!}{%
  \begin{tabular}{lc|cccc|c}
    \toprule
    \multirow{2}{*}{Models} & \multirow{2}{*}{Size} & \multicolumn{4}{c|}{Kinematic Categories} & \multirow{2}{*}{Average Score} \\
          &      & 2S & 2D & 3S & 3D & \\
    \midrule
    \rowcolor{groupgray}\multicolumn{7}{l}{\textit{Proprietary models}} \\
    Gemini-3.1 Flash~\citep{google2025gemini31flashlite} & -- & 49.4 & 47.5 & \textbf{61.4} & {61.1} & 54.8 \\
    ChatGPT-5.1~\citep{openai2025gpt51} & -- & \textbf{56.9} & 34.6 & 45.6 & 56.4 & 48.4 \\
    Gemini-2.5 Pro~\citep{comanici2025gemini25} & -- & 45.9 & 38.6 & 40.7 & 60.2 & 46.4 \\
    Gemini-2.5 Flash~\citep{comanici2025gemini25} & -- & 42.8 & 31.9 & 47.0 & 54.7 & 44.1 \\
    Grok 4.1 (Fast Reasoning)~\citep{xai2025grok41} & -- & 23.4 & 30.5 & 46.3 & 46.8 & 36.8 \\
    ChatGPT-5~\citep{openai2025gpt5} & -- & 32.2 & 24.6 & 35.6 & 38.1 & 32.6 \\
    ChatGPT-5 Pro~\citep{openai2025gpt5} & -- & 16.2 & 22.7 & 20.2 & 18.9 & 19.5 \\
    \addlinespace[2pt]
    \rowcolor{groupgray}\multicolumn{7}{l}{\textit{Open-weight models}} \\
    Qwen3-VL-32B-Instruct~\citep{bai2025qwen3} & 32B & 38.1 & 39.7 & 39.8 & 43.0 & 40.2 \\
    InternVL-3.5-30B~\citep{wang2025internvl3} & 30B & 33.1 & 33.0 & 31.4 & 44.7 & 35.5 \\
    Qwen3-VL-8B-Instruct~\citep{bai2025qwen3} & 8B & 17.2 & 27.6 & 36.0 & 48.3 & 32.3 \\
    Qwen3.5-4B~\citep{qwen3_5} & 4B & 26.6 & 35.7 & 19.8 & 41.3 & 31.2 \\
    InternVL-3.5-8B~\citep{wang2025internvl3} & 8B & 26.9 & 23.2 & 38.4 & 31.7 & 30.0 \\
    Qwen3-VL-2B-Instruct~\citep{bai2025qwen3} & 2B & 25.0 & 28.6 & 16.0 & 39.1 & 27.2 \\
    Phi-4-Multimodal-Instruct~\citep{abouelenin2025phi-4} & 5.6B & 26.6 & 26.5 & 31.6 & 23.8 & 27.1 \\
    SmolVLM-Instruct~\citep{marafioti2025smolvlm} & 0.26B & 30.0 & 21.4 & 22.8 & 33.0 & 26.8 \\ 
    Qwen3.5-2B~\citep{qwen3_5} & 2B & 28.7 & 29.7 & 17.9 & 25.9 & 25.6 \\
    InternVL-3.5-2B~\citep{wang2025internvl3} & 2B & 24.4 & 22.7 & 15.3 & 34.9 & 24.3 \\
    Molmo-7B~\citep{deitke2025molmo} & 7B & 13.8 & 20.3 & 19.1 & 41.7 & 23.7 \\
    CogVLM2 Video~\citep{hong2024cogvlm2} & 12B & 18.1 & 16.8 & 16.3 & 24.0 & 18.8 \\
    VILA-7B~\citep{lin2024vila} & 7B & 14.4 & 19.5 & 8.8 & 30.4 & 18.3 \\
    Phi-3-Mini-128K-Instruct~\citep{Abdin2024Phi3TR} & 3.8B & 14.7 & 12.4 & 21.6 & 19.6 & 17.1 \\
    MiniCPM-V 4.5~\citep{yu2025minicpm} & 8B & 27.5 & 33.2 & 0.0 & 0.0 & 15.2 \\
    LLaVA-13B~\citep{liu2023llava} & 13B & 11.2 & 10.8 & 8.1 & 21.9 & 13.0 \\
    \addlinespace[2pt]
    \rowcolor{groupgray}\multicolumn{7}{l}{\textit{Ours}} \\
    \rowcolor{ourspurple}\textbf{\method{}-VL-4B} & \textbf{4B} & 45.4 & 55.4 & 45.8 & 56.0 & 50.6 \\
    \rowcolor{ourspurple}\textbf{\method{}-VL-9B} & \textbf{9B} & 55.0 & 52.9 & {55.6} & 58.1 & 55.4 \\
    \rowcolor{ourspurple}\textbf{\method{}-VL-27B (Reasoning)} & \textbf{27B} & {48.7} & \textbf{62.4} & 60.5 & \textbf{62.8} & \textbf{58.6} \\
    \bottomrule
  \end{tabular}
  }
\end{table}

\begin{figure}[p]
    \centering
    \includegraphics[width=\linewidth]{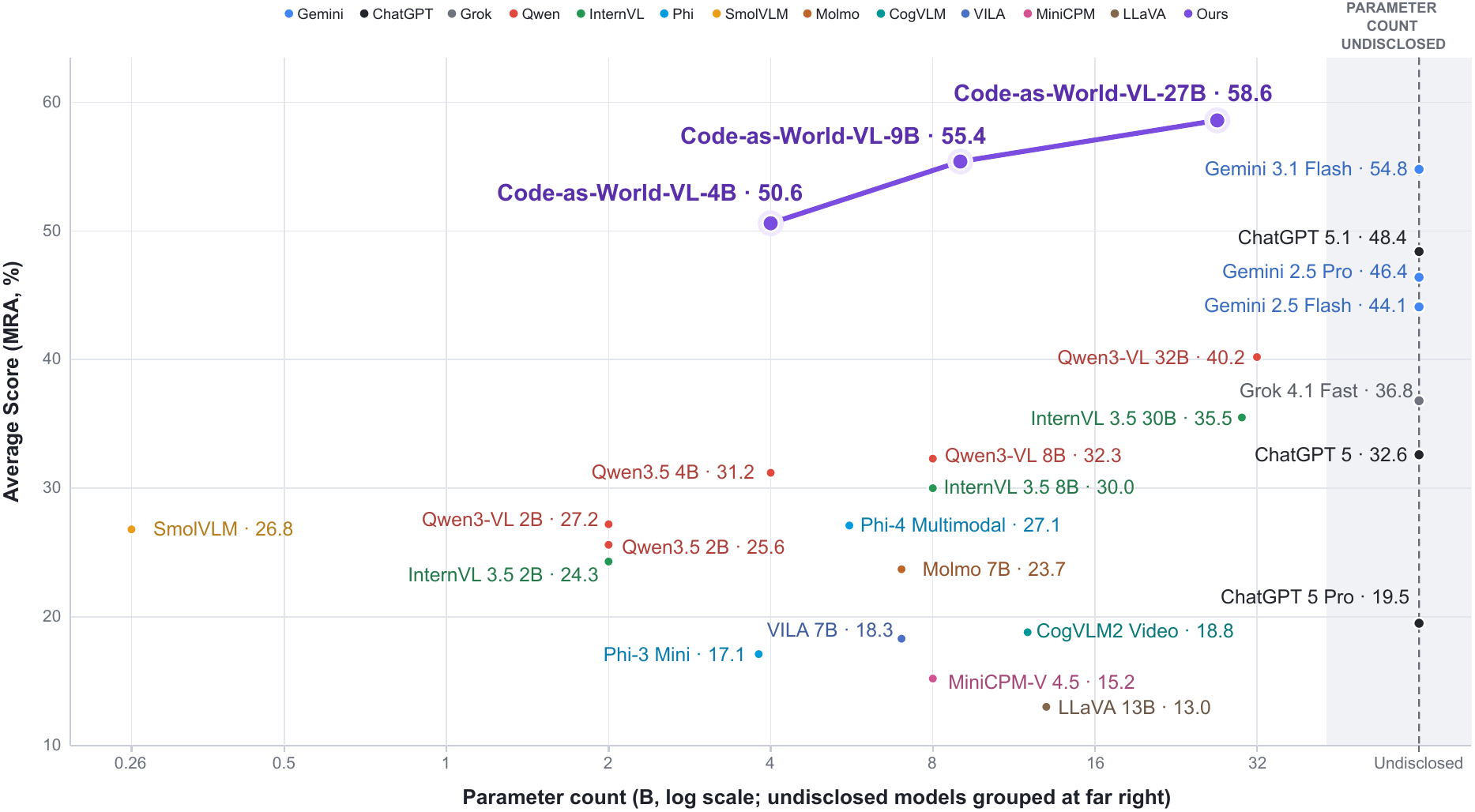}
    \caption{\textbf{Parameter scaling on QuantiPhy.} Models with
    undisclosed parameter counts are grouped at the far right.}
    \label{fig:scaling-curve}
\end{figure}

\begin{figure*}[t]
  \centering
  \definecolor{tracegray}{RGB}{246,246,246}
  \definecolor{tracegreen}{RGB}{229,243,234}
  \definecolor{traceyellow}{RGB}{252,244,220}
  \setlength{\tabcolsep}{1.5pt}
  \setlength{\fboxsep}{3pt}

  \begin{minipage}{0.98\textwidth}
    \centering
    \begin{tabular}{@{}cccc@{}}
      \includegraphics[width=0.235\textwidth]{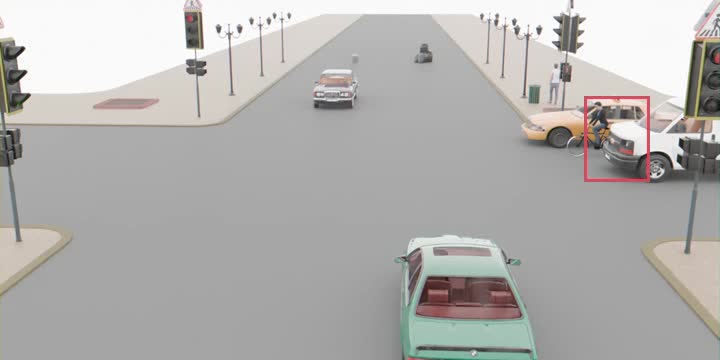} &
      \includegraphics[width=0.235\textwidth]{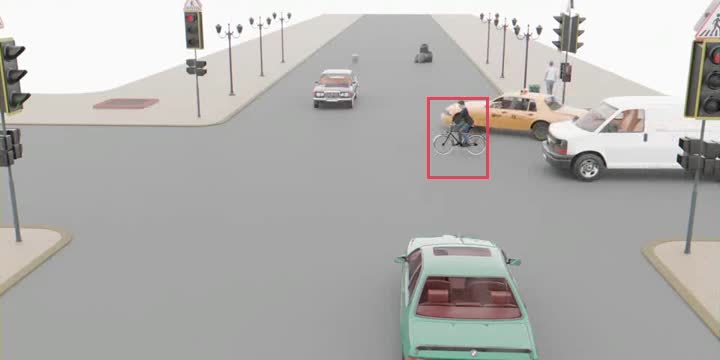} &
      \includegraphics[width=0.235\textwidth]{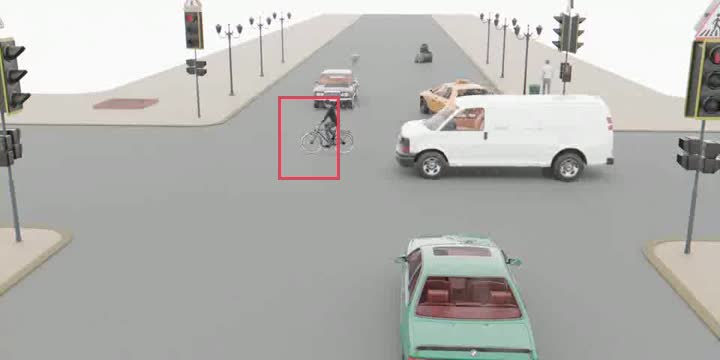} &
      \includegraphics[width=0.235\textwidth]{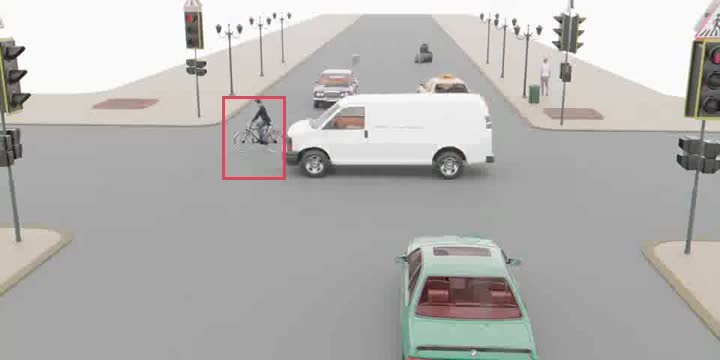} \\
      {\scriptsize $t=0.1\,\mathrm{s}$} & {\scriptsize $t=0.9\,\mathrm{s}$} &
      {\scriptsize $t=1.7\,\mathrm{s}$} & {\scriptsize $t=2.1\,\mathrm{s}$}
    \end{tabular}

    \vspace{2pt}
    \fcolorbox{black!20}{tracegray}{%
      \parbox{0.945\textwidth}{\scriptsize
      \textbf{Prior:} Bicycle-camera distances are $d=23.5534\,\mathrm{m}$ at
      $0.58\,\mathrm{s}$ and $d=23.2262\,\mathrm{m}$ at $1.58\,\mathrm{s}$. \\
      \textbf{Question:} What is the average speed of the bicycle in $\mathrm{m/s}$?}}

    \vspace{2pt}
    \fcolorbox{black!20}{traceyellow}{%
      \parbox{0.945\textwidth}{\scriptsize\raggedright
      \textbf{\texttt{<think>} (abridged).}
      \begin{enumerate}[leftmargin=1.35em,itemsep=0pt,topsep=1pt,parsep=0pt]
        \item \textbf{Identify the motion.} The bicycle moves from right to left.
        \item \textbf{Compute the radial component.}
        $v_r=|23.2262-23.5534|/(1.58-0.58)=0.33\,\mathrm{m/s}$.
        \item \textbf{Track and calibrate.} The lateral track is about $210$ pixels/s;
        mean depth $\bar z\approx23.4\,\mathrm{m}$ gives $v_{\perp}\approx5.5$--$6.8\,\mathrm{m/s}$.
        \item \textbf{Check.} Since $v_r\ll v_{\perp}$, the total speed is about $6\,\mathrm{m/s}$.
      \end{enumerate}
      \texttt{</think>}}}

    \vspace{2pt}
    \fcolorbox{black!20}{tracegreen}{%
      \parbox{0.945\textwidth}{\scriptsize
      \noindent\begin{tabular}{@{}l@{\hspace{1.5em}}l@{\hspace{1.5em}}l@{}}
        \textbf{Predicted Answer:} $6\,\mathrm{m/s}$ &
        \textbf{GT:} $5.33\,\mathrm{m/s}$ &
        \textbf{MRA:} $0.8$
      \end{tabular}}}
  \end{minipage}

  \vspace{4pt}
  {\color{black!20}\rule{0.945\textwidth}{0.35pt}}
  \vspace{3pt}

  \begin{minipage}{0.98\textwidth}
    \centering

    \begin{tabular}{@{}cccc@{}}
      \includegraphics[width=0.235\textwidth]{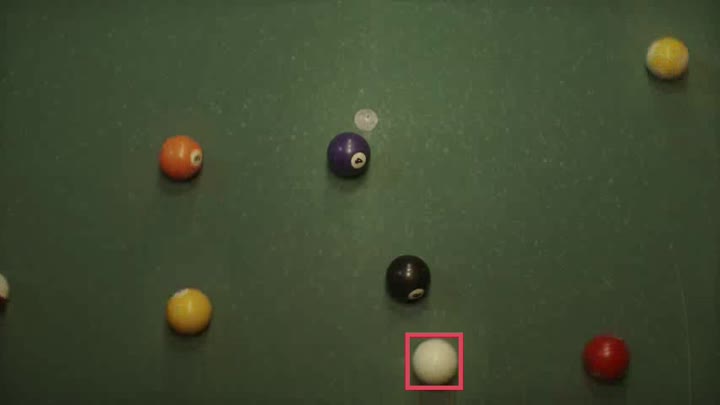} &
      \includegraphics[width=0.235\textwidth]{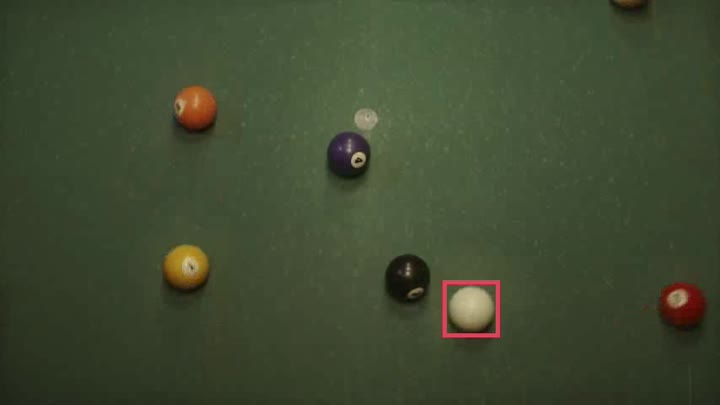} &
      \includegraphics[width=0.235\textwidth]{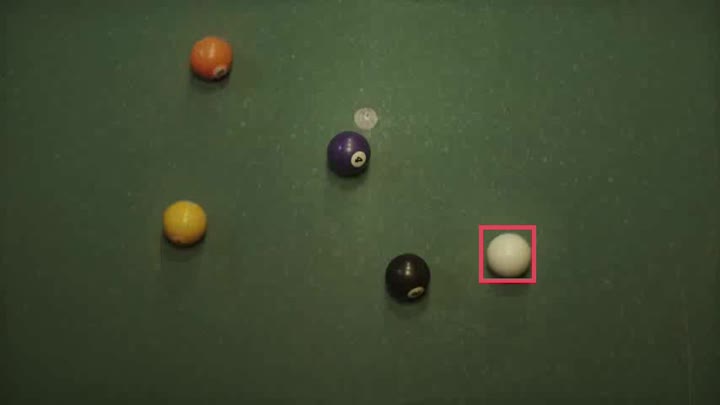} &
      \includegraphics[width=0.235\textwidth]{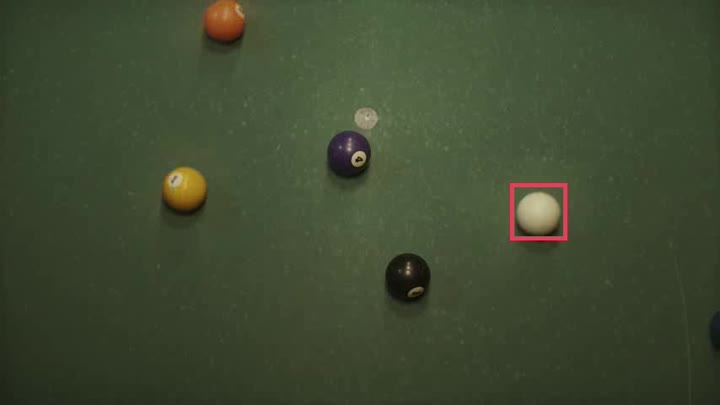} \\
      {\scriptsize $t=1.1\,\mathrm{s}$} & {\scriptsize $t=1.4\,\mathrm{s}$} &
      {\scriptsize $t=1.7\,\mathrm{s}$} & {\scriptsize $t=2.0\,\mathrm{s}$}
    \end{tabular}

    \vspace{2pt}
    \fcolorbox{black!20}{tracegray}{%
      \parbox{0.945\textwidth}{\scriptsize
      \textbf{Prior:} Billiard-ball diameter $=57.2\,\mathrm{mm}$. \\
      \textbf{Question:} What is the white ball's average velocity from $1.00$ to
      $2.00\,\mathrm{s}$ in $\mathrm{cm/s}$?}}

    \vspace{2pt}
    \fcolorbox{black!20}{traceyellow}{%
      \parbox{0.945\textwidth}{\scriptsize\raggedright
      \textbf{\texttt{<think>} (abridged).}
      \begin{enumerate}[leftmargin=1.35em,itemsep=0pt,topsep=1pt,parsep=0pt]
        \item \textbf{Locate the target.} Track the white ball (red boxes) across the table.
        \item \textbf{Measure in Image-Space.} Its center displacement over $1.00\,\mathrm{s}$
        is approximately $248$ pixels; the ball spans about $53$ pixels.
        \item \textbf{Calibrate to World-Space.} The $57.2$-mm diameter gives the local
        scale; perspective-normalized displacement yields $v\approx26.8\,\mathrm{cm/s}$.
        \item \textbf{Check.} The direction and nearly constant track agree with the observed roll.
      \end{enumerate}
      \texttt{</think>}}}

    \vspace{2pt}
    \fcolorbox{black!20}{tracegreen}{%
      \parbox{0.945\textwidth}{\scriptsize
      \noindent\begin{tabular}{@{}l@{\hspace{1.5em}}l@{\hspace{1.5em}}l@{}}
        \textbf{Predicted Answer:} $26.8\,\mathrm{cm/s}$ &
        \textbf{GT:} $26.82\,\mathrm{cm/s}$ &
        \textbf{MRA:} $1.0$
      \end{tabular}}}
  \end{minipage}

  \caption{\textbf{Reasoning process of \method{}-VL-27B on QuantiPhy.}
  Two representative cases are shown vertically: an executable-world bicycle
  example and a real-video billiards example from \texttt{internet\_0027}.
  Red boxes mark the target in the selected frames. The abridged traces preserve
  the key Image-Space measurements and World-Space calibration steps.}
  \label{fig:reasoning-trace}
\end{figure*}

\subsection{Evaluation}
\subsubsection{Experimental Setup}
\label{sec:setup}
\paragraph{Models and Training.}
We train controlled direct-answer variants of \method{}-VL at 4B and 9B using eight NVIDIA H100 GPUs and the two-phase curriculum described in Sec.~\ref{sec:training}. The first phase establishes image-space measurement through supervised fine-tuning; the second applies GRPO~\citep{guo2025deepseek} to world-level VQA derived from text-driven and video-driven executable worlds. We refer to the checkpoints after the first phase as the \emph{Image-Space} variants and to the checkpoints after both phases as \method{}-VL-4B and \method{}-VL-9B. To examine whether the framework extends to reasoning models at a larger scale, we additionally train \method{}-VL-27B (Reasoning), which produces a chain-of-thought reasoning before its final answer. During training and evaluation, all variants uniformly sample $16$ temporally ordered frames from each video. Appendix~\ref{app:reasoning-model} gives the separate 27B protocol.

\paragraph{Training and Evaluation Data.}
Our image-space supervision is constructed from four referring-expression datasets---RefCOCO~\citep{yu2016modeling}, RefCOCO+~\citep{yu2016modeling}, RefCOCOg~\citep{mao2016generation}, and RefCLEF~\cite{kazemzadeh2014referitgame}---together with GOT-10K~\cite{huang2019got}. The referring-expression datasets provide natural-language descriptions and ground-truth bounding boxes, from which we generate questions about the referred object's width, height, and diagonal length in raw pixels, as well as grounding questions requiring its bounding-box coordinates. GOT-10K provides dense object tracks over video, from which we construct questions about the target object's image-space extent, velocity, and acceleration at randomly sampled timestamps, together with timestamp-specific video grounding questions. Our world-level supervision contains text-driven and video-driven executable worlds produced by \method{}, from which we derive synchronized videos, physical states, and world-level VQA examples. We use the corresponding held-out datasets for pixel-level evaluation and the open-source QuantiPhy-validation set~\citep{puyin2026quantiphy} for metric physical reasoning evaluation.

\paragraph{Metrics.}
For image-space measurement evaluation, we evaluate on five datasets: RefCOCO, RefCOCO+, RefCOCOg, RefCLEF, and GOT-10K. For each dataset, we compute Mean Relative Accuracy (MRA)~\citep{yang2025thinking} by comparing the model's pixel-valued numerical answers with the corresponding ground-truth pixel quantities. The four referring-expression datasets evaluate object extent, while GOT-10K evaluates object extent and motion quantities. For QuantiPhy evaluation in world-space, we evaluate metric physical reasoning on the open-source QuantiPhy-validation set~\citep{puyin2026quantiphy} and compute MRA against its released ground-truth numerical answers. We additionally report the 2S, 2D, 3S, and 3D subsets under the official protocol.

\subsubsection{Image-Space Measurement}
\label{sec:perception}

Both the 4B and 9B Image-Space variants
achieve strong image-plane measurement performance despite their compact model
sizes, as shown in Table~\ref{tab:perception}. The complete \method{}-VL models further improve over their Image-Space
counterparts on every benchmark, showing that World-Space training preserves
and strengthens the underlying Image-Space grounding capability. Further
evaluation details and complete results are provided in
Appendix~\ref{app:image-space-evaluation}.

\subsubsection{Quantitative Physical Reasoning}
\label{sec:main-quantiphy}

QuantiPhy evaluates whether a model can use a physical prior to calibrate monocular video evidence into world-unit estimates of object size, displacement, velocity, and acceleration. We follow the official protocol and compare all models under the same video input and question format.

As shown in Table~\ref{tab:quantiphy-validation}, \method{}-VL delivers strong quantitative physical reasoning in the controlled direct-answer comparison. \method{}-VL-4B substantially outperforms larger open-weight baselines and remains competitive with leading proprietary systems, while \method{}-VL-9B achieves the best average performance among the direct-answer variants. Its consistent strength across the benchmark subsets shows that the gain is not confined to a particular quantity or scene configuration. Instead, \method{}-VL more reliably connects visual measurements with metric scale and motion, demonstrating the value of executable-world supervision for general quantitative reasoning from video.

To further validate this physical reasoning capability, we train a larger, reasoning-enabled \method{}-VL-27B that exposes its measurement-and-calibration reasoning trace before producing the final scalar answer. As shown in Table~\ref{tab:quantiphy-validation}, the model exceeds the direct-answer 9B variant and the strongest proprietary baseline. Because model scale and response protocol change together, we view the 27B result as evidence that the framework extends to larger reasoning models, rather than as a controlled estimate of the effect of reasoning alone. Figure~\ref{fig:reasoning-trace} illustrates this process on a QuantiPhy example, while Appendix~\ref{app:reasoning-model} details the reasoning-model setting. Figure~\ref{fig:scaling-curve} further summarizes the scaling behavior, with average MRA increasing from the 4B to 9B direct-answer variants and reaching its highest value for the reasoning-enabled 27B model.

\subsection{Limitations}

The empirical study in this section has two main limitations. First, QuantiPhy evaluates only a limited subset of physical understanding, focusing primarily on monocular scale calibration for size, displacement, velocity, and acceleration under relatively constrained motion settings. It therefore covers only a small part of real-world physics. Natural scenes can contain camera motion, rotation, deformation, occlusion, contact, collision, friction, fluids, rigid-body interactions, and long-horizon multi-object dynamics. 

Second, Code-as-World-VL currently learns from supervision derived from verified executable worlds, but does not internalize the agentic discovery process itself. In other words, the model is trained on the outcomes of world representation and verification, while hypothesis construction, simulation, diagnosis, and iterative revision remain external to the model. Extending physical reasoning to broader mechanisms and turning this discovery loop into a native model capability are important directions for future work.

\section{Related Work}

\paragraph{Physical understanding and reasoning.} Physical intelligence has been explored from increasingly richer forms of reasoning over the physical world. Spatial reasoning focuses on recovering geometric structures and relations between objects from visual observations~\citep{yang2025thinking,wu2026spatial-mllm,liu2026spatial-ttt}. Beyond spatial structure, physical question answering studies whether models can understand and reason about object properties, interactions, and measurable physical quantities from images and videos~\citep{yi2019clevrer,liu2024physics3d,chow2025physbench,zhou2025paibenchcomprehensivebenchmarkphysical,puyin2026quantiphy}. Another line of work investigates intuitive physics, evaluating whether models can acquire human-like expectations about object permanence, dynamics, and physical plausibility beyond surface-level visual correlations~\citep{riochet2018intphys,bordes2025intphys}. Increasingly, interactive physical environments have been used to study whether agents can apply physical knowledge to solve novel tasks through action and intervention~\citep{bakhtin2019phyre,xu2026deepphy,xue2023phy,wu2026perception,yao2026apple}. Despite these advances, existing approaches primarily evaluate physical understanding through task-specific outputs, while the underlying representations of physical entities, states, and mechanisms remain largely implicit.

\paragraph{Code as world representations.} Code has emerged as a promising medium for representing executable worlds. In virtual environments, recent works explore code-based world models for interactive agents, where programs serve as executable engines of environments that can be generated, modified, and evaluated through interaction~\citep{lehrach2026code,rodionov2026executable,tang2024worldcoder}. Beyond virtual worlds, researchers have also investigated code-based representations for physical content creation and simulation. At the object level, recent benchmarks study the generation of 3D assets and procedural objects through code~\citep{zheng2026voxelcodebench,gao20263dcodebench}; at the scene level, programmatic representations have been used to describe editable 3D environments and indoor scenes~\citep{zhang2025scene,wang2026scenecode,yang2026code,yin2026vision}. These approaches demonstrate the advantages of code in providing structured, compositional, and editable representations, but they primarily focus on static scene structures. More recent work extends code-driven representations to physical dynamics through executable simulations and physics-aware reasoning~\citep{liang2026visphyworld,kovavcivc2026mpmworlds,xie2026physcodebench}. However, these approaches do not address the more fundamental problem of abstracting physical mechanisms from real-world observations into executable representations.

\paragraph{Agentic optimization and discovery.}
Recent advances in agentic systems have explored how agents can autonomously discover improved solutions by iteratively proposing, evaluating, and refining external artifacts. Coding agents have demonstrated the ability to discover improved algorithms through evolutionary search and automated evaluation~\citep{novikov2025alphaevolve}, while similar ideas have been extended to automating academic experiments~\cite{autoresearch2026}, discovering reusable skills for robotics~\citep{lu2026aspire}, and generating executable representations of visual environments~\citep{yin2026vision,wang2026scenecode}. These approaches show that agents can move beyond executing predefined procedures toward actively searching for artifacts that better satisfy task objectives. Our work echoes this trend in the context of physical world representations: it treats an executable world hypothesis as the artifact to be discovered and iteratively refines it through simulation and verification against language or visual evidence.
\section{Conclusion}

Physical understanding requires more than predicting or describing observations; it requires representations that capture the mechanisms underlying how the world is composed, evolves, and can be manipulated. In this work, we introduced Code-as-World, which explores executable world representations as a bridge between visual observations and mechanism-grounded reasoning. By combining structured code representations with an evolving agentic discovery process, Code-as-World enables agents to construct, verify, and improve executable hypotheses of the observed physical world. We further showed that these verified executable worlds can provide scalable physical supervision for training vision-language models on quantitative physical reasoning. Our results suggest that executable code representations offer a promising direction toward more explicit, verifiable, and generalizable physical intelligence.

\subsection{Broader Physical Phenomena}
Our current implementation focuses primarily on rigid-body dynamics, but the Code-as-World paradigm is not tied to any particular physical regime. Code is sufficiently expressive to provide a common interface to specialized simulators for fluids \cite{Stable_fluids,dai2025rainygs}, cloth and other deformable bodies \cite{cloth_sim,narain2012adaptive}, combustion \cite{nguyen2002physically,shen2026fierygs}, fracture \cite{o1999graphical}, elasticity and plasticity \cite{terzopoulos1987elastically,irving2004invertible}, and gas dynamics \cite{fedkiw1999non}, potentially connecting decades of progress in graphics and computational physics to physical intelligence. The central challenge lies in operationalizing this expressiveness: coding agents must learn to correctly invoke and compose various simulation engines, while the discovery process must extract sufficiently informative evidence from observations and verify candidate worlds across different physical regimes. Extending both the executable representation and its evidence-verification mechanisms is therefore an important direction for future work.

\subsection{Broader Physical Capabilities}
Beyond its current use as a source of outcome supervision for quantitative physical reasoning, representing the physical world through code opens several broader directions for physical intelligence:

\begin{itemize}
    \item \textbf{General and grounded physical reasoning}: Code-as-World provides a structured representation over entities, states, relations, and dynamics, enabling supervision beyond sparse question-answering objectives. Future work can explore how executable worlds can train models to ground entities, infer physical relations, simulate possible interactions, and verify their own reasoning against explicit world states \cite{bakhtin2019phyre, lehrach2026code, rodionov2026executable}.
    \item \textbf{Physically consistent video generation}: Executable world representations provide video generators with persistent states that explicitly maintain object identity, geometry, and temporal evolution. By separating world dynamics from visual rendering \cite{alhaija2025cosmos}, future systems may achieve more coherent long-horizon generation, controllable interventions, and improved physical consistency.
    \item \textbf{Deliberate embodied interaction}: Executable world representations may support a System-2-like form of embodied intelligence, in which agents construct explicit world models, reason over possible futures, and deliberate over candidate actions before execution rather than relying solely on reactive policies \cite{intelligence2026pi}. By providing a shared abstraction across passive observations, simulated environments, and real-world interaction, they can enable physical knowledge acquired in one setting to transfer to another, offering a path toward scalable embodied learning beyond the limits of direct real-world experience.
\end{itemize}

\section*{Authors}

Hanyang Wang,
Yimo Cai,
Weiliang Chen,
Jiawei Chi,
Haowen Sun,
Qiyu Dai,
Yi-Hsin Hung,
Xingzhuo Guo,
Jinshan Ren,
Runmao Yao,
Ziwei Liu,
Mingsheng Long,
Yueqi Duan,
Jun Gao,
Jiangran Lyu,
Fangfu Liu,
Jialong Wu\corrauthor

\section*{Affiliations} MirroS, Tsinghua University, Peking University, Nanyang Technological University

\begingroup
\renewcommand{\thefootnote}{\fnsymbol{footnote}}
\footnotetext[2]{Project lead.}
\endgroup

\clearpage
\printbibliography

\clearpage
\appendix

\section{Dataset Details}
\label{app:dataset-details}

We provide additional details on the datasets summarized in
Sec.~\ref{sec:setup}. The Image-Space datasets provide direct image-plane
measurement supervision, whereas the Code-as-World executable-world dataset
provides synchronized videos, physical states, and World-Space VQA examples
from verified EWRs.

\subsection{Image-Space Datasets}
\label{app:image-space-datasets}

\paragraph{Image sources.}
We construct image-based examples from RefCOCO and
RefCOCO+~\citep{yu2016modeling}, RefCOCOg~\citep{mao2016generation}, and
RefCLEF~\citep{kazemzadeh2014referitgame}. Each annotation associates a
natural-language referring expression with a ground-truth bounding box. From
the box coordinates, we derive questions about object width, height, diagonal
extent, and relative image-plane position. We retain the original grounding
task as auxiliary supervision by requiring the model to return the box of the
referred object.

\paragraph{Video source.}
GOT-10K~\citep{huang2019got} provides dense, frame-level bounding boxes for a
tracked object. We uniformly sample 16 temporally ordered frames from each
clip and use the box centers and extents to construct questions about object
size, displacement, speed, and acceleration in pixel units. We additionally
construct timestamp-specific grounding questions from the sampled boxes. A
short semantic description of the tracked object is included in each question
to specify the target unambiguously.

\paragraph{Statistics.}
After removing samples whose images overlap with the evaluation data, the
Image-Space training set contains $73{,}335$ question--answer pairs. Of these,
$46{,}763$ are derived from GOT-10K: $20{,}858$ speed, $8{,}399$ velocity,
$12{,}169$ acceleration, $3{,}995$ grounding, and $1{,}342$ size-related
examples. The remaining $26{,}572$ examples come from the four
referring-expression datasets, including $16{,}989$ grounding and $9{,}583$
size-related examples.

\subsection{World-Space Executable-World Dataset}
\label{app:executable-world-data}

The Code-as-World executable-world dataset unifies World-Space supervision
constructed from reviewed language specifications and real-video observations.
It contains $1{,}585$ text-driven and $988$ video-driven VQA samples.

\section{Experiment Protocols}
\label{app:evaluation-protocols}

\subsection{Executable-World Fidelity and Realism Evaluation}
\label{app:ewr-evaluation}

We evaluate an executable world along three complementary axes: agreement with
the source observation, preservation of object motion, and proximity to the
distribution of real videos.

\paragraph{Visual Alignment.}
For frame $t$, let $M_t$ and $\widehat M_t$ denote the union of observed and
rendered object masks, respectively. The silhouette component is their IoU,
$s_t^{\mathrm{sil}}=|M_t\cap\widehat M_t|/|M_t\cup\widehat M_t|$, with a value
of one when both masks are empty. Let $z_t$ and $\widehat z_t$ be the observed
and rendered depth maps, and let $I_t$ and $\widehat I_t$ be their 8-bit RGB
frames. For a pixel $p$ in the mask overlap
$\Omega_t=M_t\cap\widehat M_t$, let $\Omega_t^{\mathrm{dep}}$ be the subset
with finite observed/rendered depth and positive observed depth, and let
$\Omega_t^{\mathrm{rgb}}$ be the valid RGB pixels in $\Omega_t$. We compute
the median absolute relative depth error and normalized RGB error,
\begin{equation}
  e_t^{\mathrm{dep}}
  =\underset{p\in\Omega_t^{\mathrm{dep}}}{\operatorname{median}}
   \frac{|\widehat z_t(p)-z_t(p)|}{\max(|z_t(p)|,10^{-6})},
  \qquad
  e_t^{\mathrm{rgb}}
  =\underset{p\in\Omega_t^{\mathrm{rgb}}}{\operatorname{mean}}
   \frac{\|\widehat I_t(p)-I_t(p)\|_1}{3\times255}.
  \label{eq:app-visual-errors}
\end{equation}
The corresponding similarities are
$s_t^{\mathrm{dep}}=(1+e_t^{\mathrm{dep}})^{-1}$ and
$s_t^{\mathrm{rgb}}=\max(0,1-e_t^{\mathrm{rgb}})$. Define the active component
set $\mathcal A_t$ to always contain $\mathrm{sil}$, to contain
$\mathrm{dep}$ iff $\Omega_t^{\mathrm{dep}}\ne\varnothing$, and to contain
$\mathrm{rgb}$ iff $\Omega_t^{\mathrm{rgb}}\ne\varnothing$. If $T$ is the
number of native evaluated frames, Visual Alignment is the full-video mean of
the active-weight-renormalized frame score,
\begin{equation}
  \operatorname{VA}
  =\frac{1}{T}\sum_{t=1}^{T}
   \frac{\displaystyle\sum_{c\in\mathcal A_t}w_c s_t^c}
        {\displaystyle\sum_{c\in\mathcal A_t}w_c},
   \qquad
   (w_{\mathrm{sil}},w_{\mathrm{dep}},w_{\mathrm{rgb}})
   =(0.60,0.25,0.15).
  \label{eq:app-visual-alignment}
\end{equation}
Thus, unavailable depth or RGB components are omitted before the weights are
renormalized, while the silhouette component remains active on every frame.

\paragraph{Object IoU.}
Let $M_{t,o}$ and $\widehat M_{t,o}$ be the observed and rendered masks for
object $o$ in frame $t$. We compute
\begin{equation}
  \operatorname{IoU}_{t,o}
  =\frac{|M_{t,o}\cap\widehat M_{t,o}|}
         {|M_{t,o}\cup\widehat M_{t,o}|},
  \label{eq:app-object-iou}
\end{equation}
and report Object IoU as its mean over all annotated object--frame pairs. We
set IoU to one when both masks are empty. Rendered visibility is resolved with
the final scene z-buffer so that occluded geometry is not penalized as an
additional visible region. Higher Visual Alignment and Object IoU indicate
closer visual agreement~\citep{perazzi2016benchmark}.

\paragraph{Trajectory fidelity.}
For an object, let $\mathbf p_t$ be its simulator ground-truth image-plane
center at sampled frame $t$, and let $\widehat{\mathbf p}_t$ be its CoTracker
estimate in the video being evaluated. Both the simulator render and its
sim-to-real video are initialized from $\mathbf p_0$ and compared against the
same ground-truth trajectory. For each video, we uniformly sample 16 frames
and resize it to width 512 before tracking. For a frame of width $W$ and height
$H$, let $D=\sqrt{W^2+H^2}$ be its diagonal. To keep the notation compact,
$\langle\cdot\rangle$ denotes the mean over all valid objects and sampled
frames, excluding the initialization frame; for velocity, it denotes the mean
over valid consecutive frame pairs. Following standard trajectory-error
evaluation~\citep{gupta2018social,jiang2021cotr}, we define
\begin{align}
  \operatorname{Traj\text{-}ADE}
  &=\frac{100}{D}
    \left\langle\|\widehat{\mathbf p}_t-\mathbf p_t\|_2\right\rangle,
  \label{eq:app-traj-ade}\\
  \operatorname{Velocity\text{-}ADE}
  &=\frac{100}{D}
    \left\langle
      \| (\widehat{\mathbf p}_t-\widehat{\mathbf p}_{t-1})
       -(\mathbf p_t-\mathbf p_{t-1}) \|_2
    \right\rangle,
  \label{eq:app-velocity-ade}\\
  \operatorname{Accuracy@2\%D}
  &=100\left\langle
    \mathbb{1}\!\left[
      \|\widehat{\mathbf p}_t-\mathbf p_t\|_2\leq0.02D
    \right]\right\rangle,
  \label{eq:app-accuracy-2d}
\end{align}
Traj-ADE is reported in $\%D$ and Velocity-ADE in $\%D$/step. We retain
absolute image positions and do not subtract the initial tracking offset.
Dataset-level results give each executable world equal weight; its sim-to-real
variants are averaged before the final cross-world mean.

\paragraph{JEDi.}
JEDi~\citep{luo2024jedi} measures distributional video realism using V-JEPA~\citep{bardes2024v} video features. Let
$\{\mathbf z_i^{r}\}_{i=1}^{n}$ be features from real reference videos and
$\{\mathbf z_j^{c}\}_{j=1}^{m}$ those from candidate videos. With feature
dimension $d$ and the degree-two polynomial kernel
$k(\mathbf u,\mathbf v)=(\mathbf u^\top\mathbf v/d)^2$, we report the biased
maximum mean discrepancy
\begin{equation}
  \operatorname{JEDi}
  =100\left[
    \frac{1}{n^2}\sum_{i,i'}k(\mathbf z_i^r,\mathbf z_{i'}^r)
    +\frac{1}{m^2}\sum_{j,j'}k(\mathbf z_j^c,\mathbf z_{j'}^c)
    -\frac{2}{nm}\sum_{i,j}k(\mathbf z_i^r,\mathbf z_j^c)
  \right].
  \label{eq:app-jedi}
\end{equation}
Lower JEDi indicates that the candidate video distribution is closer to the
real reference distribution.

\paragraph{TRAJAN.}
TRAJAN~\citep{allen2025direct} isolates motion realism. We extract point tracks with BootsTAPIR~\citep{doersch2024bootstap}, encode
them with the TRAJAN TrackAutoEncoder, and flatten the resulting
$128\times64$ latent into an $8192$-dimensional feature. From each feature set,
we estimate the sample mean and unbiased sample covariance $(\mu,\Sigma)$.
Given $(\mu_r,\Sigma_r)$ and $(\mu_c,\Sigma_c)$ for real and candidate videos,
the reported distance is
\begin{equation}
  \operatorname{TRAJAN}
  =\|\mu_r-\mu_c\|_2^2
   +\operatorname{Tr}\!\left(
      \Sigma_r+\Sigma_c
      -2\left(
        \Sigma_r^{1/2}
        \Sigma_c
        \Sigma_r^{1/2}
      \right)^{1/2}
    \right).
  \label{eq:app-trajan}
\end{equation}
Lower TRAJAN indicates closer agreement with real-video motion statistics.
Here $\operatorname{Tr}$ denotes the matrix trace, and each matrix square root
is the principal positive-semidefinite square root.

\subsection{Executable Engine Configurations}
\label{app:engine-configurations}

We use MuJoCo~\citep{todorov2012mujoco} as the simulation platform for
constructing and executing EWRs. Within this platform, we support two
interchangeable execution engines. The animation engine describes motion
kinematically through time-varying poses and trajectories, emphasizing what
moves and how it moves, whereas the physics engine describes mechanical bodies
and interactions, deriving motion from properties such as forces, contacts,
and constraints. Both engines implement the same executable-world interface
and can therefore be plugged into the agentic discovery process independently
of input evidence, permitting flexible evidence--engine pairings according to
the desired motion control and physical detail. The main paper illustrates the
agentic discovery process using video evidence with the animation engine,
while Section~\ref{app:sim-to-real-evaluation} reports results using video
evidence with the physics engine. After simulation, we apply a realism-oriented
re-rendering stage to the rendered videos using
Wan2.2-VACE~\citep{wan2025wan,jiang2025vace} together with an internal video
generation model. This stage conditions on the simulator render to improve
photorealism while preserving the scene structure and temporal dynamics of the
executable rollout.

\subsection{Image-Space Measurement Evaluation}
\label{app:image-space-evaluation}

The Image-Space evaluation isolates direct measurement from World-Space
calibration. For an axis-aligned box
$B_t=(x_t^{\min},y_t^{\min},x_t^{\max},y_t^{\max})$, its width, height, and
diagonal extent are
\begin{equation}
  w_t=x_t^{\max}-x_t^{\min},
  \qquad
  h_t=y_t^{\max}-y_t^{\min},
  \qquad
  \ell_t=\sqrt{w_t^2+h_t^2}.
  \label{eq:app-pixel-extent}
\end{equation}
For video, let $\mathbf c_t$ be the center of the tracked box at a uniformly
sampled timestamp with interval $\Delta t$. We compute image-plane velocity
and acceleration by central differences,
\begin{equation}
  \mathbf v_t
  =\frac{\mathbf c_{t+1}-\mathbf c_{t-1}}{2\Delta t},
  \qquad
  \mathbf a_t
  =\frac{\mathbf c_{t+1}-2\mathbf c_t+\mathbf c_{t-1}}{\Delta t^2}.
  \label{eq:app-pixel-kinematics}
\end{equation}
Scalar speed and acceleration targets are the corresponding vector magnitudes.
We generate questions only at timestamps for which all required boxes are
valid. The targets remain in raw pixels, pixels per second, or pixels per
second squared; no object-size prior or world-unit scale is supplied.

Scalar predictions are evaluated with Eq.~(\ref{eq:app-mra}), using the same
parsing and relative-error computation. We average sample-level MRA
independently within each benchmark, yielding the five dataset scores reported
in Table~\ref{tab:perception}. Grounding outputs are parsed separately as bounding
boxes; grounding examples are not included in the scalar MRA reported in the
table.

\subsection{World-Space QuantiPhy Evaluation}
\label{app:quantiphy-validation}

We evaluate on the open-source QuantiPhy validation
set~\citep{puyin2026quantiphy}. It contains 159 quantitative question--answer
pairs with visible ground-truth answers. Each example provides a monocular
video, a question, and a physical prior from which the model must infer a
scalar World-Space quantity. We follow the official categorization and
evaluation protocol.

\paragraph{Kinematic categories.}
The four reported subsets are denoted 2S, 2D, 3S, and 3D. The numeral specifies
the spatial setting: 2 denotes planar kinematics, whereas 3 denotes a
depth-aware three-dimensional setting. The letter specifies the type of the
provided source prior: S denotes a static quantity, such as object size, and D
denotes a dynamic quantity, such as velocity or acceleration. The queried
target may itself be static or dynamic; the subset is determined by the source
prior and spatial setting.

\paragraph{Mean Relative Accuracy.}
We use Mean Relative Accuracy (MRA)~\citep{yang2025thinking}. For sample $i$,
let $\widehat y_i$ denote the raw model response. The evaluator parses it as a
scalar and canonicalizes its sign as
$\widetilde y_i=|\operatorname{parse}(\widehat y_i)|$. Given ground-truth
$y_i>0$, the relative error is $r_i=|\widetilde y_i-y_i|/y_i$. Let
$\Theta=\{0.1,0.2,\ldots,0.9,0.95\}$. The sample-level score is
\begin{equation}
  \operatorname{MRA}_i
  =\frac{1}{|\Theta|}
   \sum_{\theta\in\Theta}
   \mathbb{1}\!\left[r_i<1-\theta\right].
  \label{eq:app-mra}
\end{equation}
Thus, a prediction receives credit at each of ten increasingly strict relative
error thresholds. A response that cannot be parsed as a finite number fails
all thresholds and contributes zero.

Within each of 2S, 2D, 3S, and 3D, we average the sample-level MRA values. The
overall score is the unweighted macro-average of these four subset scores,
rather than a sample-weighted average. All MRA values in the main paper are
multiplied by 100 for presentation.

\subsection{Reasoning-Model Setting}
\label{app:reasoning-model}

\paragraph{Scope and input.}
\method{}-VL-27B (Reasoning) is a separate outcome-supervised reasoning
variant, rather than a third point in the controlled 4B/9B direct-answer
comparison. It receives the same 16 temporally ordered frames, question, unit,
and benchmark-provided physical prior as the direct-answer models. For
depth-aware questions, it also receives the same benchmark-provided depth
context. It has no access to the generating EWR, simulator states, object
tracks, external tools, retrieval, or ground-truth measurements at test time.

\paragraph{Training protocol.}
The 27B variant is optimized with GRPO on the merged text-driven and
video-driven World-Space questions constructed from verified executable
worlds. Each prompt produces 16 rollouts;
the global update batch size is 16, and optimization uses AdamW with learning
rate $5\times10^{-6}$, weight decay $0.01$, five warm-up steps, and bfloat16
parameters. We enable the model's medium-effort thinking mode and impose no KL
penalty. The reward parser discards the reasoning trace and computes MRA only
from the final numerical answer. No process label or reward supervises the
content of the trace, keeping the reasoning training outcome-based.

\paragraph{Response schema.}
The system prompt supplies the opening \texttt{<think>} tag and asks the model
to close the trace with \texttt{</think>}, then emit only one scalar and the
requested unit on a single line. Within the trace, a complete solution can
identify the target and timestamps, obtain an Image-Space measurement, resolve
depth and perspective, and calibrate the measurement into a World-Space
quantity. The trace is free-form rather than a sequence of separately
supervised fields. Figure~\ref{fig:reasoning-trace} presents a representative
generated trace from the reported model.

\paragraph{Inference and parsing.}
For the result in Table~\ref{tab:quantiphy-validation}, we evaluate the model
at training step 60. We sample one response per QuantiPhy example with
temperature $1.0$, top-$p=0.95$, and a maximum response length of $6{,}144$
tokens. Subset MRA values and their unweighted macro-average in the table are
reported from this step-60 validation run. We do not use self-consistency,
majority voting, best-of-$N$ selection, or tool calls. The evaluator takes only
the text after the final \texttt{</think>} tag and extracts its first scalar; a
short one-line direct answer is accepted as a fallback, while an unclosed
multi-line reasoning trace is unparseable and receives zero. Numbers inside
the reasoning trace are thus never visible to the numerical evaluator.

\section{Additional Experiments}
\label{app:additional-experiments}

\subsection{Text-Driven Sim-to-Real Evaluation}
\label{app:sim-to-real-evaluation}

Having established that an EWR can faithfully explain its input, we next
evaluate whether it can produce videos that are both realistic and reliably
labeled in Table~\ref{tab:realism_motion}. For text-driven worlds, the simulator provides exact states,
trajectories, and physical labels, while the video generator should improve
visual realism without changing this information. We therefore evaluate
sim-to-real generation along two axes: distributional realism and motion
fidelity.

\begin{table*}[t]
    \centering
    \caption{
        Distributional realism and motion fidelity of text-driven sim-to-real
        video generation. Realism metrics use held-out real videos as the
        reference distribution, whereas motion metrics compare CoTracker
        estimates against simulator ground-truth trajectories.
    }
    \label{tab:realism_motion}
    \resizebox{\textwidth}{!}{
    \begin{tabular}{lccccc}
        \toprule
        & \multicolumn{2}{c}{\textbf{Distributional Realism}}
        & \multicolumn{3}{c}{\textbf{Motion Fidelity}} \\
        \cmidrule(lr){2-3}
        \cmidrule(lr){4-6}

        \textbf{Video Type}
        & \textbf{JEDi MMD} $\downarrow$
        & \textbf{TRAJAN Fr\'echet} $\downarrow$
        & \textbf{Traj-ADE} $\downarrow$ ($\%D$)
        & \textbf{Velocity-ADE} $\downarrow$ ($\%D$/step)
        & \textbf{Accuracy@$2\%D$} $\uparrow$ (\%) \\
        \midrule

        Simulator Render
        & 3.000
        & 406.872
        & {1.682}
        & {0.404}
        & {78.81} \\

        Sim-to-Real Video
        & {1.484}
        & {185.321}
        & 1.677
        & 0.472
        & 77.49 \\

        \bottomrule
    \end{tabular}
    }
\end{table*}

\FloatBarrier

The generated videos are closer to authentic videos in both overall
video-feature and motion-feature distributions, while retaining motion
agreement comparable to the original simulator renders. Sim-to-real
generation therefore improves visual appearance without materially changing
the physical evolution specified by the EWR. Taken together, the two
evaluations first establish that the executable representation remains
faithful to its source evidence and then show that it can produce realistic
observations without sacrificing the reliability of simulator-derived states
and labels.

\FloatBarrier

\subsection{Additional Agentic Discovery Loop}
\label{app:additional-agentic-discovery-loop}

\begin{figure}[htb]
  \centering
  \includegraphics[width=\textwidth]{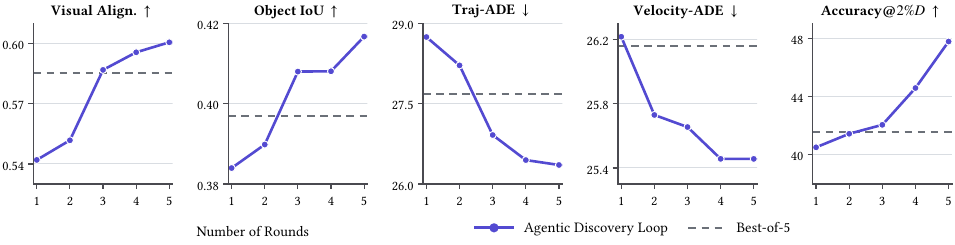}
  \caption{\textbf{Agentic discovery with the physics engine.} Solid curves
  report performance over five iterative discovery rounds, while dashed gray
  lines mark Best-of-$5$ under the same five-evaluation budget. Higher is
  better for Visual Alignment, Object IoU, and Accuracy@$2\%D$; lower is better
  for Traj-ADE and Velocity-ADE.}
  \label{fig:physics-engine-agentic-ablation}
\end{figure}

To assess whether iterative discovery remains effective beyond the animation
engine used in the main analysis, we repeat the experiment with the physics
engine described in Appendix~\ref{app:engine-configurations}. As shown in
Figure~\ref{fig:physics-engine-agentic-ablation}, performance improves
consistently over the five discovery rounds: Visual Alignment, Object IoU, and
Accuracy@$2\%D$ increase, while Traj-ADE and Velocity-ADE decrease. By the
fifth round, the agentic discovery loop also outperforms the matched-budget
Best-of-$5$ baseline on all five metrics, showing that its iterative gains
persist under a different executable engine.

\FloatBarrier

\subsection{Image-Space Measurement Results}
\label{app:image-space-results}

\begin{figure}[h]
  \centering
  \includegraphics[width=\textwidth]{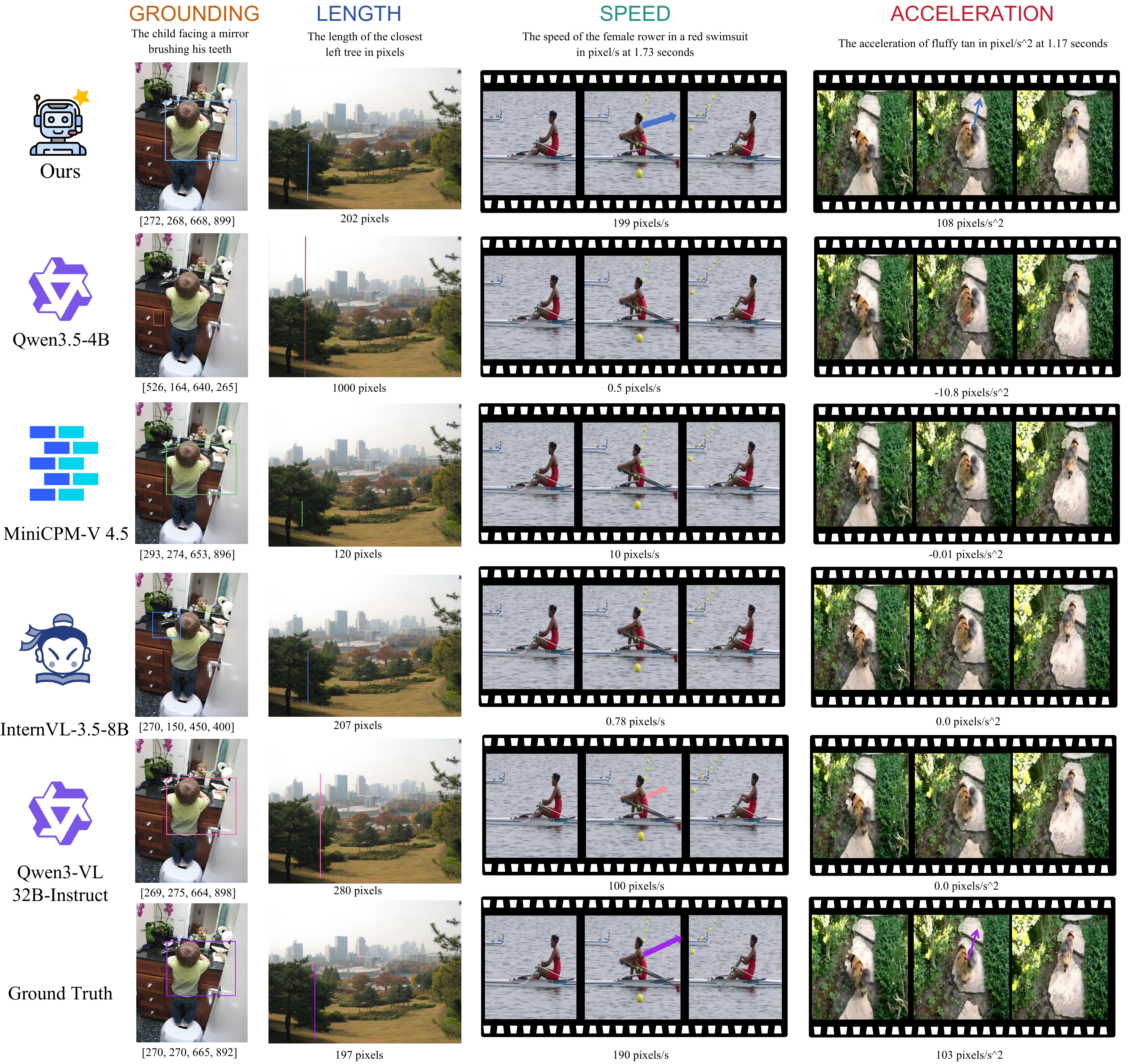}
  \caption{Qualitative comparison of Image-Space grounding and measurement.
  The four columns evaluate target grounding, object length, speed, and
  acceleration, respectively. All numerical quantities are expressed in the
  pixel-space units specified by the question.}
  \label{fig:app-perception-comparison}
\end{figure}

We compare \method{}-VL with open-weight models on four referring-expression
datasets and GOT-10K. These benchmarks cover object extent and grounding in
images as well as object extent and motion in video.
Table~\ref{tab:perception} reports the quantitative comparison. Figure~\ref{fig:app-perception-comparison} complements the quantitative results
in Table~\ref{tab:perception} with representative predictions for grounding,
length, speed, and acceleration. In the examples shown, \method-VL identifies
the intended target and produces measurements close to the ground truth,
whereas the comparison models exhibit larger localization or numerical errors.

\begin{table}[H]
  \centering
  \definecolor{ourspurple}{RGB}{234,233,249}
  \definecolor{groupgray}{RGB}{245,245,245}
  \caption{\textbf{Pixel-level measurement evaluation.} Referring-expression datasets evaluate object extent in images, while GOT-10K evaluates object extent and motion in video. Image-Space variants use only the first training phase; full \method{}-VL models additionally receive executable-world reinforcement learning.}
  \label{tab:perception}
  \resizebox{\textwidth}{!}{
  \begin{tabular}{lc|ccccc}
    \toprule
    Models & Size & RefCOCO$^{\dag}$ & RefCOCOg$^{\dag}$ & RefCOCO+$^{\dag}$ & RefCLEF$^{\dag}$ & GOT-10K\textsuperscript{*} \\
    \midrule
    \rowcolor{groupgray}\multicolumn{7}{l}{\textit{Open-weight models}} \\
    Qwen3-VL-32B-Instruct~\citep{bai2025qwen3} & 32B & 49.1 & 40.3 & 44.9 & 32.5  & 15.8 \\
    Qwen3.5-27B~\citep{qwen3_5} & 27B & 47.0 & 44.8 & 46.8 & 31.6 & 6.1 \\
    Qwen3.5-9B~\citep{qwen3_5} & 9B & 38.6 & 33.0 & 38.6 & 37.8 & 15.9 \\
    MiniCPM-V 4.5~\citep{yu2025minicpm} & 8B & 55.6 & 53.2 & 54.1 & 38.6 & 15.9 \\
    InternVL-3.5-8B~\citep{wang2025internvl3} & 8B & 36.4 & 30.9 & 35.2 & 24.2 & 14.4 \\
    Qwen3-VL-8B-Instruct~\citep{bai2025qwen3} & 8B & 43.4 & 43.0 & 41.8 & 40.0 & 18.9 \\
    Qwen3.5-4B~\citep{qwen3_5} & 4B & 28.6 & 26.1 & 30.4 & 20.9 & 3.0 \\
    InternVL-3.5-4B~\citep{wang2025internvl3} & 4B & 36.9 & 32.5 & 35.9 & 27.5 & 8.4 \\
    \addlinespace[2pt]
    \rowcolor{groupgray}\multicolumn{7}{l}{\textit{Ours}} \\
    \textbf{\method{}-VL-4B (Image-Space)} & \textbf{4B} & 56.1 & 51.6 & 56.2 & 31.2 & 18.7 \\
    \textbf{\method{}-VL-4B} & \textbf{4B} & 62.9 & 60.2 & 63.3 & 39.8 & 20.5 \\
    \textbf{\method{}-VL-9B (Image-Space)} & \textbf{9B} & 63.7 & 61.1 & 61.5 & 47.2 & 20.1 \\
    \textbf{\method{}-VL-9B} & \textbf{9B} & \textbf{68.3} & \textbf{65.6} & \textbf{66.4} & \textbf{61.9} & \textbf{26.6} \\
    \bottomrule
  \end{tabular}
  }
\end{table}

The Image-Space variants establish reliable visual grounding and pixel-level
measurement, which provide the perceptual basis for subsequent World-Space
learning. After adding World-Space supervision, the full 4B and 9B models
improve on all five Image-Space benchmarks over their corresponding
Image-Space variants. The two training stages therefore reinforce one another:
Image-Space supervision grounds physical reasoning in observable measurements,
while World-Space supervision feeds physical consistency back into grounding
and quantitative measurement.

\FloatBarrier

\subsection{Ablation of Data Sources}
\label{app:data-ablation}

We ablate the three data sources in our two-phase curriculum: Image-Space
measurement data $\mathcal D_{\mathrm{pix}}$, text-driven executable worlds
$\mathcal D_{\mathrm{text}}$, and video-driven executable worlds
$\mathcal D_{\mathrm{video}}$. We perform this ablation on our compact 4B and
9B models. All variants first use
$\mathcal D_{\mathrm{pix}}$ to establish Image-Space measurement, after which
the two World-Space sources are introduced separately or jointly.

\begin{table}[H]
  \centering
  \scriptsize
  \definecolor{ablationpurple}{RGB}{234,233,249}
  \caption{\textbf{Ablation of data sources.} We report MRA on the QuantiPhy
  subsets and their macro-average for the three data sources in our two-phase
  curriculum.}
  \label{tab:ablation-sources}
  \resizebox{\textwidth}{!}{%
  \begin{tabular}{lc|ccc|ccccc}
    \toprule
    Training Variants & Size & $\mathcal D_{\mathrm{pix}}$ & $\mathcal D_{\mathrm{text}}$ & $\mathcal D_{\mathrm{video}}$ & 2S & 2D & 3S & 3D & Avg. \\
    \midrule
    Image-Space Only & 4B & \ensuremath{\checkmark} &  &  & 45.3 & 45.1 & 31.6 & 54.6 & 44.2 \\
    + Text-Driven Worlds & 4B & \ensuremath{\checkmark} & \ensuremath{\checkmark} &  & 47.5 & 49.4 & 40.0 & 57.0 & 48.5 \\
    + Video-Driven Worlds & 4B & \ensuremath{\checkmark} &  & \ensuremath{\checkmark} & 42.2 & 52.7 & 46.5 & 49.8 & 47.8 \\
    \rowcolor{ablationpurple}
    \textbf{Full \method{}-VL} & 4B & \ensuremath{\checkmark} & \ensuremath{\checkmark} & \ensuremath{\checkmark} & 45.4 & \textbf{55.4} & 45.8 & 56.0 & 50.6 \\
    \midrule
    Image-Space Only & 9B & \ensuremath{\checkmark} &  &  & 44.3 & 51.9 & 51.4 & 60.0 & 50.9 \\
    + Text-Driven Worlds & 9B & \ensuremath{\checkmark} & \ensuremath{\checkmark} &  & 47.2 & 48.1 & 53.5 & 61.1 & 52.5 \\
    + Video-Driven Worlds & 9B & \ensuremath{\checkmark} &  & \ensuremath{\checkmark} & 43.1 & 50.5 & \textbf{57.4} & \textbf{61.5} & 53.1 \\
    \rowcolor{ablationpurple}
    \textbf{Full \method{}-VL} & 9B & \ensuremath{\checkmark} & \ensuremath{\checkmark} & \ensuremath{\checkmark} & \textbf{55.0} & {52.9} & {55.6} & 58.1 & \textbf{55.4} \\
    \bottomrule
  \end{tabular}%
  }
\end{table}

As shown in Table~\ref{tab:ablation-sources}, adding either World-Space source
improves the 4B Image-Space model, and combining both sources achieves the best
average of $50.6$, demonstrating their complementary benefits. The same trend
holds at the larger scale, where the full 9B model improves from $50.9$ to
$55.4$. Together, these results show that exact simulator supervision and
real-video alignment contribute complementary signals beyond Image-Space
grounding.

\FloatBarrier

\begin{figure}[h]
  \centering
  \includegraphics[width=\textwidth]{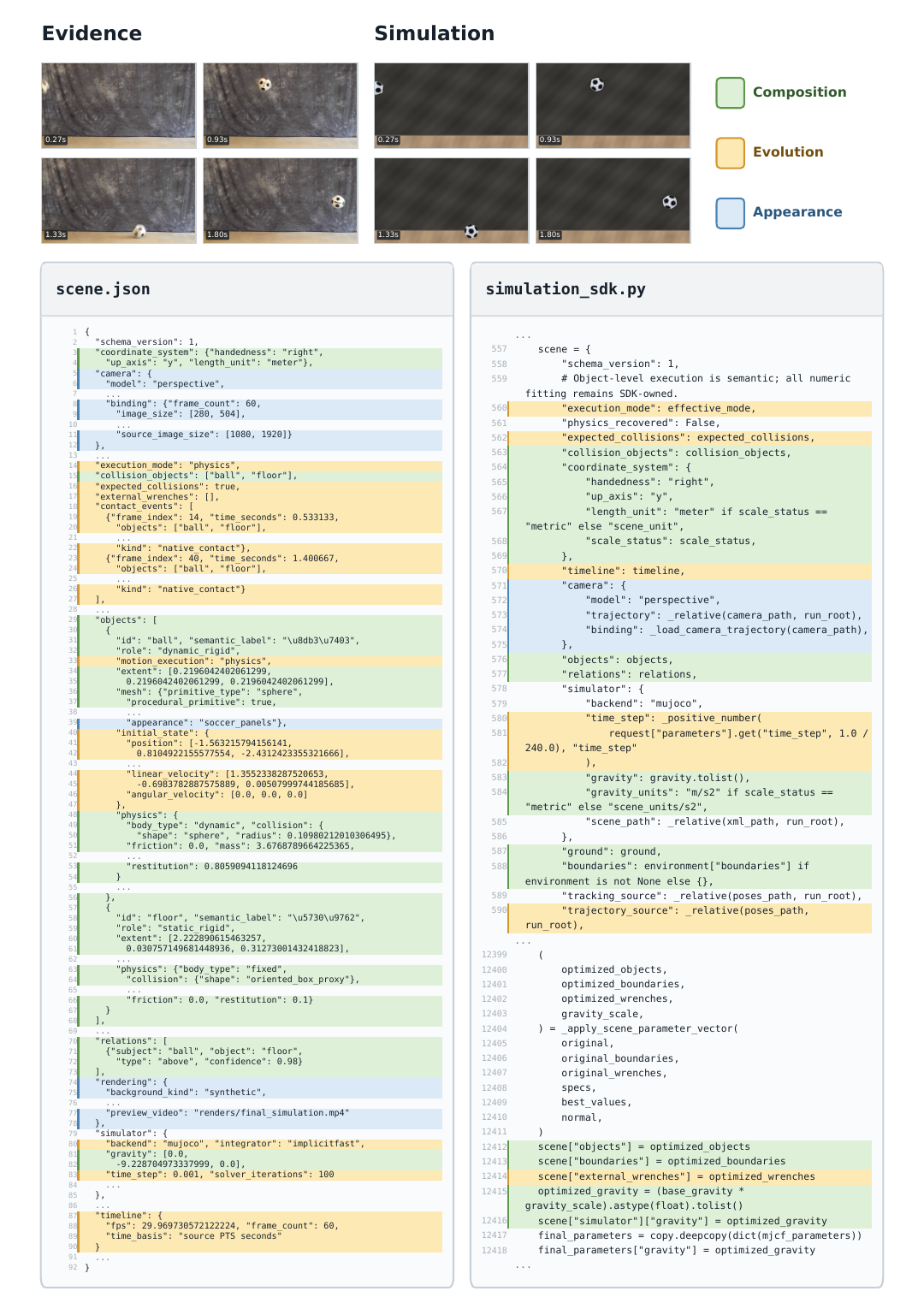}
  \caption{\textbf{A concrete EWR code and simulation interface.} A
  video-driven scene is encoded as a structured \texttt{scene.json}
  specification of its composition, evolution, appearance, physical
  parameters, relations, simulator configuration, and timeline. The
  simulator SDK instantiates and executes this specification in MuJoCo, while
  the frame-aligned evidence and simulation outputs above illustrate the
  resulting reconstruction.}
  \label{fig:scene-representation}
\end{figure}

\end{document}